\documentclass[10pt,journal,compsoc]{IEEEtran}
\usepackage{cite}
\usepackage{amsmath,amssymb,amsfonts}
\usepackage{graphicx}
\usepackage{textcomp}
\usepackage{subfigure}

\usepackage{caption}
\usepackage{mathtools}
\usepackage{array}
\usepackage{mdwmath}
\usepackage{mdwtab}
\usepackage{multirow}
\usepackage{textcomp}
\usepackage{url}
\usepackage{booktabs}
\usepackage{float}
\usepackage{enumerate}
\usepackage{xcolor,epstopdf} 
\usepackage{colortbl,adjustbox,cases}
\usepackage[square,sort,comma,numbers]{natbib}
\usepackage{algorithm} 
\usepackage{algorithmic}  
\usepackage[algo2e]{algorithm2e} 

\ifCLASSINFOpdf
\else
\fi
\begin{document}
	%
	\title{Time-aware Graph Embedding: A temporal smoothness and task-oriented approach}
	\author{Yonghui~Xu, Shengjie~Sun, Yuan~Miao, Dong~Yang, Xiaonan~Meng, \\Yi~Hu, Ke~Wang, Hengjie~Song, Chuanyan~Miao*}
	
	\author{Yonghui~Xu, Shengjie~Sun, Yuan~Miao, Dong~Yang, Xiaonan~Meng, \\Yi~Hu, Ke~Wang, Hengjie~Song, and Chuanyan~Miao*
		\IEEEcompsocitemizethanks{
			\IEEEcompsocthanksitem Yonghui~Xu, and Chuanyan~Miao are with the Nanyang Technological University in Singapore. E-mail:\{ASCYMiao\}@ntu.edu.sg. Shengjie~Sun, Dong~Yang, Xiaonan~Meng, and Yi~Hu are with the Alibaba Group.Yuan~Miao is with the Victoria University, Australia. Ke~Wang is with the Simon Fraser University, Canada. Hengjie~Song is with the South China University of Technology, China.
		}
	}
	
	%
	%

	\markboth{Journal of \LaTeX\ Class Files,~Vol.~1, No.~1, July~2020}%
	{Shell \MakeLowercase{\textit{et al.}}: Bare Demo of IEEEtran.cls for Computer Society Journals}
	%



	\IEEEtitleabstractindextext{%
		\begin{abstract}
			Knowledge graph embedding, which aims to learn the low-dimensional representations of entities and relationships, has attracted considerable research efforts recently. However, most knowledge graph embedding methods focus on the structural relationships in fixed triples while ignoring the temporal information. Currently, existing time-aware graph embedding methods only focus on the factual plausibility, while ignoring the temporal smoothness which models the interactions between a fact and its contexts, and thus can capture fine-granularity temporal relationships. This leads to the limited performance of embedding related applications. To solve this problem, this paper presents a Robustly Time-aware Graph Embedding (RTGE) method by incorporating temporal smoothness. Two major innovations of our paper are presented here. At first, RTGE integrates a measure of temporal smoothness in the learning process of the time-aware graph embedding. Via the proposed additional smoothing factor, RTGE can preserve both structural information and evolutionary patterns of a given graph. Secondly, RTGE provides a general task-oriented negative sampling strategy associated with temporally-aware information, which further improves the adaptive ability of the proposed algorithm and plays an essential role in obtaining superior performance in various tasks. Extensive experiments conducted on multiple benchmark tasks show that RTGE can increase performance in entity/relationship/temporal scoping prediction tasks.
		\end{abstract}
		
		}

	\maketitle

	\IEEEdisplaynontitleabstractindextext

	%
	\IEEEpeerreviewmaketitle

	\IEEEraisesectionheading{\section{Introduction}\label{sec:introduction}}
	With the rapid growth of Knowledge Graph (KG) construction, YAGO~\citep{suchanek2007yago}, Wiki~\citep{Leblay:2018:DVT:3184558.3191639}, Freebase~\citep{bollacker2008freebase}, DBpedia~\citep{lehmann2015dbpedia}, NELL~\citep{carlson2010toward} and many other Knowledge Bases (KB) have been created for many real-world applications, \emph{i.e.,} semantic parsing~\citep{berant2013semantic}, named entity disambiguation~\citep{damljanovic2012named}, information extraction~\citep{daiber2013improving}, recommender systems~\citep{yin2019deeper} and question answering~\citep{zhang2018variational}.
	In order to provide an effective and efficient way to solve the graph analytics problem in these applications, \emph{i.e.,} relation extraction~\citep{weston2013connecting}, entity classification~\cite{wang2017knowledge}, link prediction~\cite{du2019joint}, entity resolution~\citep{nickel2011three}, and Graph Embedding (GE) methods~\cite{guan2019knowledge} have been proposed. The key idea of GE is to map components (\emph{i.e.,} head entity, tail entity, and relation of the triple $<head-entity, relation, tail-entity>$) or sub-graph of a KG onto a low dimensional space in which the graph information is preserved. By representing a sub-graph (or head entity, tail entity, relation) as a low dimensional space vector, graph analytics can be conducted accurately and efficiently.
	
	Current research on knowledge graph embedding~\citep{cui2018survey,goyal2018graph,cai2018comprehensive,Huang:2019:GRN:3292500.3330941} has mainly concerned graphs with fixed triples~\citep{bordes2013translating}. However, in real-life scenarios, graphs, like social graphs in Twitter, citation graphs in DBLP~\citep{yang2015defining}, are time-variant, and many relations are only valid for a certain period of time. In these applications, the underlying graph structure keeps on changing continuously over time. \emph{i.e.,} in social graphs~\cite{li2018influence}, when a new user registers onto the graph or friendship is established between two users, a new entity/relation appear in the graph. However, when a user cancels the account or a friendship breaks down, previously established entity/relation disappears. The entity/relation representation of the users should be updated accordingly, such that the learned entity/relation can reflect the temporal evolution of their social relationship. Similarly, due to the frequent publication of a new graph that cites existing technology, the citation graph~\cite{lauscher2018linked} of scientific papers is continuously enriched. As a result, the influence of the article, and sometimes even the classification, has changed over time. The node embedding needs to be updated to reflect this change. In financial networks~\cite{gai2019networks}, transactions are naturally timestamped. If a user is a victim of credit card fraud, or the user's account is involved in money laundering, the characteristics of the user's account may change due to the nature of the transaction involved. In these cases, early detection of such changes is critical to improving law enforcement efficiency and reducing financial institution losses. These unique characteristics of the time-aware graph make traditional graph embedding methods fail to work since static graph embedding methods completely ignore the time-varying information of the graph and cannot capture the evolutionary patterns of the given time-aware graph. Therefore, how to design an embedding method to models the interactions between a fact and its contexts for a time-aware graph is critical. 
	
	To embed the temporal information~\cite{goyal2020dyngraph2vec} of the graph in the learning model while maintaining the inherent structural information of the given time-aware graph, an obvious way is to slice the graph into different time bins~\citep{dasgupta2018hyte}. Then the embeddings can be learned on these bins separately. Although these kinds of models take into account the temporal information of the graph in the embedding process, they cannot explicitly model the temporally-aware information. This is because these kinds of models are fit on different time bins independently, and cannot share statistical strength between two adjacent time bins. What is worse, such a model trained independently in a fixed time bin cannot remain robust when the structure of the graph changes drastically at a specific time. Suffering from the defects explained above, the current research work on temporal embedding is rather limited. It is necessary to design a graph embedding algorithm that can fit the current graph structure well, and simultaneously it does not deviate too dramatically from recent history.
	
	Previous studies~\cite{bordes2013translating,dasgupta2018hyte} about graph embedding always use the same sampling strategy for different specific tasks, \emph{i.e., head/tail entity prediction task, relationship prediction task.} 
	\begin{itemize}
		\item The (head/tail) entity prediction task: we observe the relation and (tail/head) entity in the triple $<head-entity, relation, tail-entity>$, and predict the head/tail entity. 
		\item The relationship prediction task: we observe the head entity and tail entity in the triple $<head-entity, relation, tail-entity>$, and predict the relation according to the learned model. 
	\end{itemize}
	The triples in the knowledge graph represent real facts and can only be used as positive samples in the process of training the model. To achieve better performance in different specific tasks, we need to construct negative samples based on the triples in the knowledge graph. Most existing sampling strategies obtain negative samples by replacing the head or tail entities of the triples. This kind of negative sampling strategy can obviously improve the performance of the model in the (head/tail) entity prediction task. However, these strategies would not explicitly replace relationships in the triples. This may cause positive and negative samples for the relationship to be unbalanced. Potentially imbalanced data may cause the model learned by the embedding algorithm to be unfavorable for relational prediction tasks. Furthermore, errors introduced by relational embeddings may be passed on to head or tail entity embeddings. This makes graph embedding suffer from the limitations associated with a biased sampling strategy. As a result, how to design a sampling strategy for specific tasks is crucial for graph embedding.
	
	To address the aforementioned issues, we propose a robustly time-aware graph embedding algorithm to encode temporal information in the learned embeddings directly. Particularly, RTGE slices the temporally-scoped input knowledge graph into multiple static subgraphs in which each subgraph corresponds to a timestamp. And then RTGE projects the entities and the relations of each subgraph onto temporally aware hyperplanes. We define a temporal smoothness between hyperplanes of adjacent time steps. By maintaining the temporal smoothness, we expect RTGE can avoid the hyperplanes deviate too dramatically from recent history. Moreover, we propose a task-oriented negative sampling strategy. By performing negative sampling in a balanced manner for both entities and relationships, we hope to obtain training triples with balanced positive and negative samples of entities and relationships. So that the learned entity embedding and relationship embedding can be applied to a variety of head entity/tail entity/relation prediction tasks. We highlight our contributions as follows:
	\begin{itemize}
		\item Different from previous time-aware graph embedding methods (\emph{i.e.,} t-TransE, HyTE), which learn hyperplanes of adjacent time steps independently,  RTGE attempts to maintain the temporal smoothness between hyperplanes of adjacent time steps. Thus, RTGE can model the evolution of KGs more accurately and obtain better performance in the applications.
		
		\item Unlike the existing graph embedding algorithms which only perform negative sampling on entities in different tasks, we designed a task-oriented negative sampling strategy in time-aware graph embedding, which can do negative sampling for both entities and relationships in different tasks. The newly proposed strategy can well avoid the problem of sampling data imbalance caused by biased sampling. 
	\end{itemize}
	
	We organized the rest of this paper as follows. In the next section, we review the related work. Section~\ref{sec:model} presents the problem and our proposed method and how we format the method into an optimization problem in detail. Experimental results on benchmark datasets are reported in Section~\ref{sec:exp}. Finally, Section~\ref{sec:conclusion} concludes this paper and discuss future work.

	\section{Related Works}\label{sec:related work}
	Knowledge graph embedding has been an active research area for the past couple of years. Various graph embedding methods~\citep{liao2019lanczosnet} have been put forward. Among the existing GE methods, approaches related to our study can be summarized into three categories: \emph{static graph embedding}, \emph{dynamic/incremental graph embedding}, and \emph{time-aware graph embedding}. 
	
	\subsection{Static Graph Embedding Methods}
	As a static graph embedding method, TransE~\citep{bordes2013translating} proposes an energy-based model for entities embeddings, by requiring the tail entity embedding to be close to the head entity embedding plus a vector corresponding to the relationship. Despite TransE is efficient and straightforward, it has flaws in dealing with reflexive / one-to-many/many-to-one / many-to-many relations. Different from TransE~\citep{wang2014knowledge}, TransH models a relation as a hyperplane together with a translation operation on it. By utilizing the relation-specific hyperplanes, TransH overcomes the flaws of TransE in dealing with reflexive / one-to-many/many-to-one / many-to-many relations. Unlike TransE and TransH which put both entities and relations within the same semantic space, TransR~\citep{lin2015learning} build entity and relation embeddings in separate entity space and relation spaces. In this way, TransR can avoid the problem of insufficient common space in modeling. TransD~\citep{ji2015knowledge} as an improvement of TransR proposes representing a named symbol object (entity and relation) by two vectors. The first vector represents an entity (or relation), the other vector is used to construct a mapping matrix for each entity-relation pair. By considering the diversity of entities and relations in the process of the construction mapping matrix, TransD encodes more discriminative information and obtains better results than TransR. 
	
	Recently, much effort has been invested in neural embedding for a static graph. For instance, DistMult~\citep{YangYHGD14a} presents a general neural network framework for multi-relational representation learning. In particular, By utilizing bilinear objective for relation representations, DistMult captures compositional semantics of relations. Besides, DistMult successfully extracts Horn rules that involve compositional reasoning. Instead of using embeddings containing real numbers as DistMult, ComplEx~\citep{trouillon2016complex} discuss and demonstrate the capabilities of complex embeddings. In this way, ComplEx well avoids overfitting problems caused by the explosion of parameters in existing embedded models when dealing with symmetrical/antisymmetric relationships. DistMult and ComplEx learn less powerful features than deep, multi-layer models since they only focus on shallow, fast models that can scale to large graphs. To solve the problem of shallow architectures, and the overfitting problem of fully connected deep architectures, ConvE~\citep{dettmers2018conve} proposes to use parameter efficient, fast operators which can be composed into deep networks. Recently, more and more TransE-based or neural network-based graph embedding algorithms (\emph{i.e.,}TranSparse~\citep{ji2016knowledge},  TransF~\citep{feng2016knowledge}) are proposed for static graph embedding, and they generate effective results on static graph data sets. However, none of the above methods try to combine the temporal information to explore the evolutionary pattern of the knowledge graph. 
	
	\subsection{Dynamic / Incremental Graph Embedding Methods}
	Dynamic/incremental KG embedding aims to learn embedding in an online fashion when the KG is frequently updated. For instance,~\cite{tay2017non} proposes puTransE (Parallel Universe TransE), an online and robust adaptation of TransE. To capture the temporal information for each edge in the knowledge graphs, ~\cite{trivedi2017know} presents a novel deep evolutionary knowledge network, Know-Evolve which learns nonlinearly evolving entity representations over time. Based on Generalized SVD Decomposition and matrix perturbation theory, ~\cite{zhu2018high} dynamically updates the node representation of the dynamic network while retaining high-order similarity. When the network structure changes at the next moment, ~\cite{zhu2018high} can quickly and effectively update the representation of the node. Recently,~\cite{pareja2019evolvegcn}~extends graph convolutional network, GCN~\cite{kipf2016semi} to the dynamic setting by utilizing a recurrent mechanism to update the parameters. In this way, \cite{pareja2019evolvegcn} expects to capture the dynamism of the graphs. Unlike the dynamic graph neural network algorithms which require to retrain a model or wait for convergence, \cite{liu2019real} develops new approaches to the problems of streaming graph embedding, by only updating the representations of a small proportion of vertices. In this way, \cite{liu2019real} has low space and time complexity to generate latent representations for new vertices under specified iteration rounds. ~\cite{singer2019node} proposes to initialize node embeddings with respect to the static graph. Then the initial node embeddings are aligned at different time points and eventually adapted for the specific task with a joint optimization. To model complex and nonlinearly evolving dynamic processes of the dynamic graph, ~\citep{trivedi2019dyrep} proposes a deep temporal point process model based on specially designed temporally attentive representation network. By this method, ~\citep{trivedi2019dyrep} learn to encode structural-temporal information over the graph into low dimensional representations.
	
	\subsection{Time-aware graph Embedding Methods}
	The method proposed in this paper, RTGE, is a typical time-aware graph embedding method. Different from dynamic/incremental graph embedding, time-aware graph embedding, which tries to learn the evolving patterns of a graph and incorporate time information into embedding learning and learns embedding in an offline fashion. For instance, based on TransE, t-TransE~\citep{jiang2016encoding} provides a link prediction method by using temporal order constraints to model transformation between time-sensitive relations. In the embedding process, t-TransE enforces the embeddings to be temporally consistent. Similarly, HolE~\citep{nickel2016holographic} earlier attempts to consider such temporal information for graph embedding. To incorporate the valid time of facts, ~\cite{jiang2016towards} proposes a time-aware graph embedding approach with a joint time-aware inference model using temporal consistency information as constraints. In this way, ~\cite{jiang2016towards} expects to be more accurate concerning various temporal constraints. Unlike translation based embedding methods, for example, TransE, TransH, and TransR, which ignores the time information of the graph and learns the embedding representation by defining a global margin-based loss function over the data. ~\cite{jia2017knowledge} proposes to encode temporal information of the graph by adaptively adjusting the optimal margin over time. 
	
	The most related work to our study is the hyperplane-based temporally aware knowledge graph embedding method (HyTE) proposed by  ~\citep{dasgupta2018hyte}. By investigating different hyperplanes to represent different time (i.e., segregate the embedding space into different time zones by these hyperplanes), HyTE attempts to encode temporal information directly in the learned embeddings. Note that, although the basic ideas behind HyTE and our proposed RTGE are similar, i.e., to learn hyperplanes for different time bins, RTGE differs from HyTE in two aspects: 1) In HyTE, the hyperplanes of adjacent time intervals are independent of each other. On the contrary, in RTGE, we introduce the concept of timing smoothing to optimize and learn the hyperplanes of adjacent time intervals jointly. In this way, RTGE can avoid the problem of missing timing associations between embedded spaces caused by independent learning of hyperplanes of adjacent time intervals. 2) HyTE only uses a negative sampling strategy based on randomly replacing the head or tail entities in the negative sampling process. RTGE adds relation-based negative sampling on the basis of HyTE. In this way, RTGE can avoid the problem of an imbalance of positive and negative samples due to the lack of relationship-based negative sampling.

	\section{Time-aware graph Embedding Model}\label{sec:model}
	\subsection{Problem Statement}
	Let $\mathcal{G}=\{<\widehat{h_i}, \widehat{\ell_i}, \widehat{\zeta_i}, t_s^i,t_e^i>|1\leq i\leq N\}$ denotes a time-aware graph, where $<\widehat{h_i}, \widehat{\ell_i}, \widehat{\zeta_i}>$ indicates the triple of the graph. $\widehat{h_i}$ denotes a head entity, $\widehat{\ell_i}$ denotes a tail entity and $\widehat{\zeta_i}$ denotes a relation. 
	$t_s^i$ and $t_e^i$ indicate the start and end timestamps of the fact represented by triplet $<\widehat{h_i},\widehat{\ell_i},\widehat{\zeta_i}>$, respectively. $t_s^i$ is the start timestamps of the fact, and $t_e^i$ is the end timestamps of the fact. 
	For series of given timestamp, $t\in{1,2,\dots,T}$, the time-aware graph $\mathcal{G}$ can be split into multiple static graphs $\mathcal{G}_1...\mathcal{G}_T$,  and each graph consists of several triples that are valid in the corresponding timestamp, e.g., knowledge graph $\mathcal{G}$ can be denoted by, 
	\begin{equation}
	\mathcal{G}=\mathcal{G}_1 \cup \mathcal{G}_2 \cup\dots \mathcal{G}_t \dots \cup \mathcal{G}_T
	\end{equation}
	where $t\in [1,T]$.
	Here, we denote the embeddings of head entity $\widehat{h_i}$, tail entity $\widehat{\zeta_i}$ and relation $\widehat{\ell_i}$ by $h_i$, $\zeta_i$ and $\ell_i$, respectively. Time-aware graph embedding aims to learn $h_i \in \mathcal{R}^{d\times 1}$, $\zeta_i \in \mathcal{R}^{d\times 1}$ and $\ell_i \in \mathcal{R}^{d\times 1}$ for each entity and relation, and the appropriate mapping functions $\Gamma_t, t\in [1,T]$ to meet the requirements of the following three tasks.
	\begin{itemize}
		\item Head prediction task: for an incomplete fact $<?, \widehat{\ell_i}, \widehat{\zeta_i}>$ at $t$, $\Gamma_t$ can predict the head entity $\widehat{h_i}$.
		
		\item Relation prediction task: for an incomplete fact $<\widehat{h_i}, ?, \widehat{\zeta_i}>$ at $t$, $\Gamma_t$ can predict the relation $\widehat{\ell_i}$.
		
		\item Tail prediction task: for an incomplete fact $<\widehat{h_i}, \widehat{\ell_i}, ?>$ at $t$, $\Gamma_t$ can predict the tail entity $\widehat{\zeta_i}$.
		
		\item Temporal scoping prediction task: for a fact $<\widehat{h_i}, \widehat{\ell_i}, \widehat{\zeta_i}>$ , $\Gamma_t$ can predict the temporal scoping of this fact.
		
	\end{itemize}
	
	\begin{figure*}[t!]
		\centering
		{\includegraphics[width=0.7\textwidth]{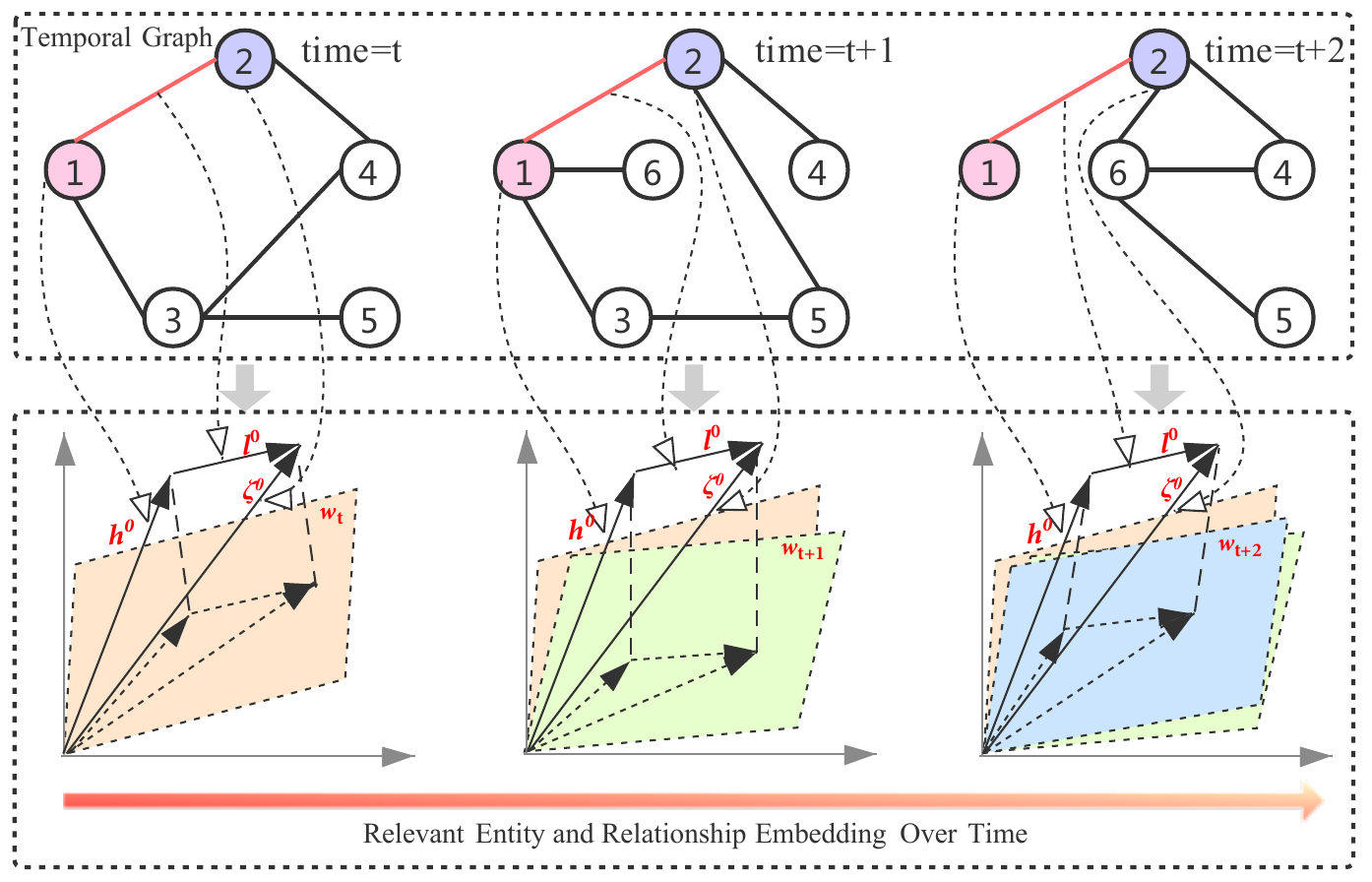}}
		\caption{Illustration of time-aware graph and its relevant embedding  evolution over time.}\label{model:fig}
	\end{figure*}
	\subsection{Modeling Embeddings Over Time}
	This section deduces how temporal information can be integrated into graph embedding based on TransE. 
	The basic idea of TransE is that it models relationships as translations in the embedding space, and enfores the embedding of the tail entity $\widehat{\zeta_i}$ to be close to the embedding of the head entity $\widehat{h_i}$ plus some vector that depends on the relationship $\widehat{\ell_i}$, (\emph{i.e.,} $h_i + \ell_i \approx \zeta_i$) when the triplet $<\widehat{h_i}, \widehat{\ell_i}, \widehat{\zeta_i}>$ holds. Obviously, if the triplet $<\widehat{h_i}, \widehat{\ell_i}, \widehat{\zeta_i}>$ does not hold, TransE enfores $h_i + \ell_i$ to be far away from $\zeta_i$. Based on this idea, TransE can obtain the embeddings by minimizing a margin-based ranking loss over the whole training set~\citep{bordes2013translating}.
	
	TransE provides a basic framework to embed entity and relation in the semantic space. Though it works well on irreflexive and one-to-one relations for a static graph, it has problems to deal with reflexive or many-to-one/one-to-many/many-to-many relations for a time-aware graph. For instance, fact: "\textit{Jone lives in Beijing in 2018}", can be described with a triple as (\ref{eq:tri2018}).
	\begin{equation} \label{eq:tri2018}
	<\widehat{h_i}:John, \widehat{\ell_i}:lives-in, \widehat{\zeta_i}:Beijing, t_s^i:2018, t_e^i:2018>
	\end{equation}
	\begin{equation} \label{eq:tri2019}
	<\widehat{h_j}:John, \widehat{\ell_j}:lives-in, \widehat{\zeta_j}:Singapore, t_s^j:2019, t_e^j:2019>
	\end{equation}
	For another fact: "\textit{Jone lives in Singapore in 2019}", we can describe it as (\ref{eq:tri2019}).
	Since: \textit{"Jone moved from Beijing to Singapore in 2019"}, the triple (\ref{eq:tri2018}) and the triple (\ref{eq:tri2019}) have the same head entity and relationship but has different tail entities. If the time information is not considered, TransE mandates $h_i + \ell_i = \zeta_i$ and $h_j + \ell_j = \zeta_j$. Since $h_i=h_j$ and $\ell_i=\ell_j$, TransE will deduce a wrong conclusion $\zeta_i=\zeta_j$. 
	
	In order to avoid the above problem, temporal aware hyperplane can be utilized to segregate the embedding space into different time zones. With the help of temporal aware hyperplane at time $t$, the representation of the triple valid at time $t$ will be projected onto hyperplane $w_t\in \mathcal{R}^{d\times 1}$ as follows,
	\begin{equation} \label{eq:qh}
	\mathcal{Q}_t(h_i)=h_i-(w_t^\top h_i)w_t
	\end{equation}
	\begin{equation} \label{eq:qr}
	\mathcal{Q}_t(\ell_i)=\ell_i-(w_t^\top \ell_i) w_t
	\end{equation}
	\begin{equation} \label{eq:qt}
	\mathcal{Q}_t(\zeta_i)=\zeta_i-(w_t^\top \zeta_i) w_t
	\end{equation}
	where $\mathcal{Q}_t(h_i)$, $\mathcal{Q}_t(\ell_i)$ and $\mathcal{Q}_t(\zeta_i)$ denote the projection of the head entity $\widehat{h_i}$, the relationship $\widehat{\ell_i}$, and the tail entity $\widehat{\zeta_i}$ on the hyperplane $w_t$, respectively.
	With this approach, triples with the same head entity and the same relationship at different times will be projected into different subspaces, and the tail entities of these triples will be represented as different embeddings in different subspaces. Thus, the many-to-one problem caused by time is avoided.
	
	Following the strategy adopted in previous method, HyTE, we learn the embeddings in (\ref{eq:qh}), (\ref{eq:qr}) and (\ref{eq:qt}) by  minimizing a margin-based ranking loss over the training set,
	\begin{eqnarray}\label{eq:obj_hyte}
	&&\arg\min_{W,h,\ell,\zeta}g(W,h,\ell,\zeta)=\\ \label{eq:obj_hyte2}
	&&\sum_{t=1}^{T}\sum_{s_i^+}^{S_t^+}\sum_{s_j^-}^{S_t^-}
	\max{(\mathcal{L}(s_i^+)+\gamma-\mathcal{L}(s_j^-),0)} \nonumber
	\end{eqnarray}
	where $W=[w_1,w_2,\dots,w_{T}]$. $s_i^+$ indicates the fact $<\widehat{h_i}, \widehat{\ell_i}, \widehat{\zeta_i}>$ which is valid during timestamp $t$. $s_i^+$ can be denoted as,
	\begin{equation}\label{eq:obj:s+}
	s_i^+=<h_i,\ell_i,\zeta_i> \in S_t^+.
	\end{equation}
	$s_j^-$ indicates the negative fact which is not valid during timestamp $t$.  $s_j^-$ can be generated by replacing the head entity or the tail entity of a valid fact  $<\widehat{h_j}, \widehat{\ell_j}, \widehat{\zeta_j}>$.
	If we use a negative head entity, $s_j^-$ can be defined as,
	\begin{equation}\label{eq:obj:sh-}
	s_j^-=<h_j^{'},\ell_j,\zeta_j> \in S_t^-.
	\end{equation}
	For a negative tail entity, $s_j^-$ can be defined as,
	\begin{equation}\label{eq:obj:st-}
	s_j^-=<h_j,\ell_j,\zeta_j^{'}> \in S_t^-.
	\end{equation}
	$\mathcal{L}$ in (\ref{eq:obj_hyte}) indicates the loss corresponding to the projection of triples on $w_t$. For instance, 
	\begin{equation} \label{eq:loss-l}
	\mathcal{L}(s_i^+)=\mathcal{L}(h_i,\ell_i,\zeta_i)=\|\mathcal{Q}_t(h_i)+\mathcal{Q}_t(\ell_i)-\mathcal{Q}_t(\zeta_i)\|
	\end{equation}
	
	\subsection{Temporal Smoothness}
	According to the introduction in the previous section, the optimal $W$, $h$, $\ell$, $\zeta$ can be solved by minimizing (\ref{eq:obj_hyte}). However, as can be seen from (\ref{eq:obj_hyte2}), the model (\ref{eq:obj_hyte}) is only the total sum of the margin-based ranking criterion on each subgraph. For the fixed $h$, $\ell$, $\zeta$, the $w_t$ learning process corresponding to each timestamp is independent of other timestamps. This result in the model learned by minimizing (\ref{eq:obj_hyte}) only containing subgraph structure information at different timestamps, but not the evolutionary pattern of the global graph.
	
	To explain this problem more clearly, Fig.~\ref{model:fig} shows an illustration of a time-aware graph and its relevant embedding evolution over time. As the figure shows, $h^0$, $\zeta^0$, and $\ell^0$ indicates the embedding of node 1, node 2, and the associated relationship at time $0$, respectively. $\omega_t$, $\omega_{t+1}$ and $\omega_{t+2}$ represent the optimal hyperplane at time $t$, $t+1$ and $t+2$, respectively. 
	From the figure, we can see that there are differences in the optimal hyperplanes at different times, and there is a close relationship between the optimal hyperplanes at adjacent times. This example validates the necessity of introducing temporal smoothness in the time-aware graph embedding model.  
	
	Actually, a graph may not change much in a short period of time, so the embedded space should not change too much. Inspired by this factor, we propose constraining the variation between hyperplanes at adjacent timestamps while accumulating the margin-based ranking criterion of each subgraph. Therefore, we define the temporal smoothness by minimizing the Euclidean distance between hyperplanes in adjacent timestamps. Formally, the corresponding loss function is, 
	\begin{equation}\label{eq:obj:smooth}
	g(W)_{smooth}=\sum_{t=1}^{T-1}\|w_{t+1}-w_t\|_2
	\end{equation}
	
	The smoothness for RTGE defined in (\ref{eq:obj:smooth}) aims to request the coordinates of an entity or a relationship to be very close to the average coordinates of its neighbors over adjacent time. For example, a person may experience the following four events at different moments,
	
	\textit{\textbf{Born}} ($t_1$)$\longrightarrow$
	\textit{\textbf{Go to school}} ($t_2$)$\longrightarrow$
	\textit{\textbf{Graduate from a school}} ($t_3$)$\longrightarrow$
	\textit{\textbf{Go to a company to work}} ($t_4$)$\longrightarrow$
	\textit{\textbf{Death}} ($t_5$)
	
	The four events mentioned above naturally have a time sequence. For example, "\textit{Go to school}" cannot occur before "\textit{Born}". "\textit{Go to a company to work}" cannot happen after "\textit{Death}". When learning the representation of $t_1 \dots t_5$, if the constraints of temporal smoothness are lacking, $t_1$ may be farther away from $t_2$ and $t_3$, but closer to $t_5$. This may cause the model to observe "\textit{Go to a company to work}" at the previous moment and predict "\textit{Born}" at the next moment. This prediction is clearly contrary to the facts. Therefore, it is necessary to keep the timing smooth constraint in the process of learning time embedding.

	\subsection{Task-Oriented Negative Sampling}
	Another problem with the model (\ref{eq:obj_hyte}) occurs during the negative sampling process. This section analyzes the advantages and disadvantages of existing negative sampling strategies in different tasks, and proposes a task-oriented negative sampling strategy based on this analysis. Note that (\ref{eq:obj_hyte}) can obtain a large number of negative sample triples by randomly replacing the head or tail entities in the training triples. By introducing a negative sample into the loss function in (\ref{eq:obj_hyte}), the discriminating ability of the model concerning the head or tail entity can be significantly improved, thereby improving the predicted performance of the learned embeddings for head or tail entity prediction task. 
	
	By comparing (\ref{eq:obj:s+}), (\ref{eq:obj:sh-}) and (\ref{eq:obj:st-}), we can find that the negative sampling of the relationship is not shown in the negative sampling strategy of (\ref{eq:obj_hyte}). Although some negative samples about relationships can be obtained in negative entity sampling, in the absence of independent relationship negative sampling, the balance of positive and negative samples is difficult to guarantee. In this case, the performance of the model (\ref{eq:obj_hyte}) in relational prediction tasks is limited. This problem is even more serious when the number of relationships is large, or the relationship has a high degree of similarity. 
	
	Based on the above analysis, the existing negative sampling strategy is more suitable for entity prediction tasks, but not suitable for relational prediction tasks. To improve the robustness and adaptability of graph embedding algorithms in different tasks, we propose a task-oriented negative sampling strategy that considers the negative sampling of the relationship while considering the negative sampling of the head and tail entities. The specific loss function and the sampling method are as follows,
	\begin{eqnarray} \label{eq:obj:task}
	&& g_{task}(W,h,\ell,\zeta)=\\
	&&\sum_{t=1}^{T}\sum_{s_i^+}^{S_t^+}
	\sum_{s_{j,e}^-,s_{k,r}^-}^{S_{t,e}^-,S_{t,r}^-}
	\max{[2\mathcal{L}(s_i^+)+
		\gamma-\mathcal{L}(s_{j,e}^-)-\beta\mathcal{L}(s_{k,r}^-),0]} \nonumber
	\end{eqnarray}
	where $g_{task}(W,h,\ell,\zeta)$ is the loss function corresponding to specific task. $S_t^+$ indicates the positive triple set for training,
	\begin{eqnarray}\label{eq:st+}\nonumber
	&&S_t^+=\{<h_i,\ell_i,\zeta_i>|\\
	&&<h_i,\ell_i,\zeta_i>\in \mathcal{G}_t, i\in[1,|\mathcal{G}_t|]\} 
	\end{eqnarray}
	$S_{t,e}^-$ indicates the sampled negative triple set for training which is generated by replacing the head or tail entity in $S_t^+$,
	\begin{eqnarray}\label{eq:ste-} \nonumber 
	&&S_{t,e}^-=\{<h_i^{'},\ell_i,\zeta_i>,<h_i,\ell_i,\zeta_i^{'}>|\\\nonumber 
	&&<h_i,\ell_i,\zeta_i>\in \mathcal{G}_t,  
	<h_i^{'},\ell_i,\zeta_i>\in \overline{\mathcal{G}_t},\\
	&& <h_i,\ell_i,\zeta_i^{'}>\in \overline{\mathcal{G}_t}\} 
	\end{eqnarray}
	$S_{t,r}^-$ indicates the sampled negative triple set for training which is generated by replacing the relation in $S_t^+$,
	\begin{eqnarray}\label{eq:str-}\nonumber
	&&S_{t,r}^-=\{<h_i,\ell_i^{'},\zeta_i>|\\ 
	&&<h_i,\ell_i,\zeta_i>\in \mathcal{G}_t, 
	<h_i,\ell_i^{'},\zeta_i>\in \overline{\mathcal{G}_t}\}
	\end{eqnarray}
	$|\mathcal{G}_t|$ indicates the size of $\mathcal{G}_t$.
	$\overline{\mathcal{G}_t}$ denotes the complement of $\mathcal{G}_t$, $\overline{\mathcal{G}_t} \cup \mathcal{G}_t = \mathcal{G}$ and $\overline{\mathcal{G}_t} \cap \mathcal{G}_t = \emptyset$.
	
	$\mathcal{L}(s_{j,e}^-)$ in (\ref{eq:obj:task}) represents the loss function associated with the entity negative triple, and $\mathcal{L}(s_{k,r}^-)$ represents the loss function associated with the relationship negative triple. Compared to (\ref{eq:obj_hyte2}), (\ref{eq:obj:task}) replaces the loss $\mathcal{L}(s_j^-)$ associated with the head entity and the tail entity negative triple in (\ref{eq:obj_hyte2}) with a linear combination of $\mathcal{L}(s_{j,e}^-)$ and $\mathcal{L}(s_{k,r}^-)$. Following consideration of the difference between the magnitude of the entity and the relationship, excessive use of negative triples about relationships may affect the ability of the model to discriminate on the entity. To avoid this problem, we add a parameter $\beta$ for $\mathcal{L}(s_{k,r}^-)$ to adjust its weight. In this way, we introduce the discriminant characteristics of the relationship in the model (\ref{eq:obj:task}), so that the model can be applied to both the prediction task of the entity and the prediction task of relationship.
	
	\subsection{Model Learning}\label{sec:opt}
	By combining (\ref{eq:obj:smooth}) and (\ref{eq:obj:task}), we derive an overall optimization approach for the following unified optimization problem, which is constructed by using $\mathcal{J}(W,h,\ell,\zeta)$ as a general loss function, 
	\begin{eqnarray}\label{eq:obj}
	&&\arg\min_{W,h,\ell,\zeta} \mathcal{J}(W,h,\ell,\zeta)=\\ \nonumber
	&&\alpha* g_{smooth}(W)+g_{task}(W,h,\ell,\zeta)\\ \nonumber
	&s.t.& \|h_i\|_2=1, \|\ell_i\|_2=1, \|\zeta_i\|_2=1, i\in [1,N].\\ \nonumber
	&& \|w_t\|_2=1, t\in [1,T].
	\end{eqnarray}
	where $\alpha$ is a tradeoff parameter. 
	In this section, we derive approaches to solve the optimization problems constructed in (\ref{eq:obj}). Firstly, we convert the optimization problem to an unconstrained one,
	\begin{eqnarray}\label{eq:obj-nons}
	&&\arg\min_{W,h,\ell,\zeta} \mathcal{J}(W,h,\ell,\zeta)=\\ \nonumber
	&&\alpha* g_{smooth}(W)+g_{task}(W,h,\ell,\zeta) + \\ \nonumber
	&& \xi\sum_{t=1}^{T}(\|w_t\|_2-1)^2+ \xi\sum_{i=1}^{N}[(\|h_i\|_2-1)^2 \\ \nonumber
	&&+ (\|\ell_i\|_2-1)^2 + (\|\zeta_i\|_2-1)^2],
	\end{eqnarray}
	where $\xi$ is a tradeoff parameter.
	Then, we propose an alternating optimization algorithm to learn $W$, $h$, $\ell$ and $\zeta$ alternatively and iteratively. To be specific, at the $\rho$-th iteration, we first fix the matrix $h$, $\ell$ and $\zeta$ and update the value of each $w_t$ in $W$ using gradient descent based on the following rule,
	\begin{eqnarray}\label{eq:obj-u-w}
	(w_t)_{\rho+1}=(w_t)_\rho - \psi \frac{\partial \mathcal{J}(W,h,\ell,\zeta)}{\partial w_t}
	\end{eqnarray}
	where $\psi$ indicates the learning rate, and
	\begin{eqnarray}\label{eq:obj-w}\nonumber
	&& \frac{\partial \mathcal{J}}{\partial w_t}= \alpha(\frac{w_t-w_{t+1}}{\|w_{t+1}-w_t\|_2}+\frac{w_t-w_{t-1}}{\|w_t-w_{t-1}\|_2}) \\\nonumber
	&& +\sum_{s_i^+}^{S_t^+}
	\sum_{s_{j,e}^-,s_{k,r}^-}^{S_{t,e}^-,S_{t,r}^-}
	[2
	\nabla_{w_t}{\mathcal{L}(s_i^+)} -    
	\nabla_{w_t}{\mathcal{L}(s_{j,e}^-)} \\\nonumber
	&&-\beta
	\nabla_{w_t}{\mathcal{L}(s_{k,r}^-)} ]_\dagger  + 2\xi(\|w_t\|_2-1)w_t,
	\end{eqnarray}
	where $[ \dots ]_\dagger$ in (\ref{eq:obj-w}) is an indication function. If $2{\mathcal{L}(s_i^+)}-{\mathcal{L}(s_{j,e}^-)} -\beta {\mathcal{L}(s_{k,r}^-)}>0$ then $[x]_\dagger=x$, otherwise $[x]_\dagger=0$.
	
	After updating the value of $W$, we then alternatively fix $W$ and update $h$, $\ell$ and $\zeta$ respectively, based on the following rule,
	\begin{eqnarray}\label{eq:obj-u-h}
	(h_u)_{\rho+1}=(h_u)_\rho - \psi \frac{\partial \mathcal{J}(W,h,\ell,\zeta)}{\partial h_u}
	\end{eqnarray}
	\begin{eqnarray}\label{eq:obj-u-r}
	(\ell_u)_{\rho+1}=(\ell_u)_\rho - \psi \frac{\partial \mathcal{J}(W,h,\ell,\zeta)}{\partial \ell_u}
	\end{eqnarray}
	\begin{eqnarray}\label{eq:obj-u-t}
	(\zeta_u)_{\rho+1}=(\zeta_u)_\rho - \psi \frac{\partial \mathcal{J}(W,h,\ell,\zeta)}{\partial \zeta_u}
	\end{eqnarray}
	where
	\begin{eqnarray}\label{eq:obj-h}\nonumber
	\frac{\partial \mathcal{J}}{\partial h_u}=\sum_{t=1}^{T}\sum_{s_i^+}^{S_t^+}
	\sum_{s_{j,e}^-,s_{k,r}^-}^{S_{t,e}^-,S_{t,r}^-}
	[2
	\nabla_{h_u}{\mathcal{L}(s_i^+)} -    
	\nabla_{h_u}{\mathcal{L}(s_{j,e}^-)}\\ \nonumber
	- \beta \nabla_{h_u}{\mathcal{L}(s_{k,r}^-)} ]_\dagger + 
	2\xi(\|h_u\|_2-1)h_u,
	\end{eqnarray}
	
	\begin{eqnarray}\label{eq:obj-r}\nonumber
	\frac{\partial \mathcal{J}}{\partial \ell_u}=\sum_{t=1}^{T}\sum_{s_i^+}^{S_t^+}
	\sum_{s_{j,e}^-,s_{k,r}^-}^{S_{t,e}^-,S_{t,r}^-}
	[2
	\nabla_{\ell_u}{\mathcal{L}(s_i^+)} -    
	\nabla_{\ell_u}{\mathcal{L}(s_{j,e}^-)}\\ \nonumber
	- \beta \nabla_{\ell_u}{\mathcal{L}(s_{k,r}^-)} ]_\dagger + 2\xi(\|\ell_u\|_2-1)\ell_u,
	\end{eqnarray}
	
	\begin{eqnarray}\label{eq:obj-t}\nonumber
	\frac{\partial \mathcal{J}}{\partial \zeta_u}=\sum_{t=1}^{T}\sum_{s_i^+}^{S_t^+}
	\sum_{s_{j,e}^-,s_{k,r}^-}^{S_{t,e}^-,S_{t,r}^-}
	[2
	\nabla_{\zeta_u}{\mathcal{L}(s_i^+)} -    
	\nabla_{\zeta_u}{\mathcal{L}(s_{j,e}^-)}\\ \nonumber
	- \beta \nabla_{\zeta_u}{\mathcal{L}(s_{k,r}^-)} ]_\dagger + 2\xi(\|\zeta_u\|_2-1)\zeta_u.
	\end{eqnarray}
	
	We alternatingly and iteratively update $W$, $h$, $\ell$ and $\zeta$ until the change in values of the objective function $\mathcal{J}$ is less than a threshold $\epsilon$. 
	Considering that the possible combination of negative triples is enormous, we sample several negative triples for each training triplet. The negative triples are randomly sampled and include three groups, ($M$ negative head entity samples, $M$ negative relationship samples, and $M$ negative tail entity samples). 
	
	\begin{algorithm}[t]
		\SetAlgoLined
		\KwResult{Output $W$, $h$, $\ell$ and $\zeta$}
		Initialization $W$, $h$, $\ell$ and $\zeta$, maximum number of iterations $\kappa$, threshold $\epsilon$.
		
		\While{$t \leq T$}{
			Sampling entity negative tripe set $S_t^+$ based on (\ref{eq:ste-}).
			Sampling relation negative tripe set $S_r^+$ based on (\ref{eq:str-}).
		}
		\While{the number of iteration $\leq$ $\kappa$ or $\mathcal{J}(W,h,\ell,\zeta)$ in (\ref{eq:obj}) does not converge}{
			\If{$W$ does not converge}{
				Update the value of each $w_t$ in $W$ by (~\ref{eq:obj-u-w}).
			}
			\If{$h$, $\ell$ or $\zeta$ do not converge}{
				Update head entity embedding $h$ by (\ref{eq:obj-u-h}).\\
				Update relation embedding $\ell$ by (\ref{eq:obj-u-r}).\\
				Update tail entity embedding $\zeta$ by (\ref{eq:obj-u-t}).\\
			}
		}
		\caption{RTGE}\label{alg:opt}
	\end{algorithm}
	
	\subsection{Computational Complexity}
	In this section, we analyze the computational complexity of our proposed RTGE. 
	The total time spent is mainly determined by the time complexity of computing $\mathcal{J}$'s gradients concerning $W$, $h$, $\ell$ and $\zeta$.
	The computational cost for computing the gradient of $\mathcal{J}$ with respect $W$, $h$, $\ell$ and $\zeta$ is $O(CT\kappa d)$, where $C$ indicates the number of constraint pairs consisting of positive and negative sample triples, and $d$ is the dimensionality of each embedding. Moreover, the computational cost caused by other operations, in Algorithm~\ref{alg:opt} is not more than $O(CT\kappa d)$. Therefore, the overall computational complexity of Algorithm~\ref{alg:opt} is $O(CT\kappa d)$.

	\renewcommand\arraystretch{1.2}
	\begin{table} 
		\centering
		\caption{Characteristic information of the Wikidata12K Data set and YAGO11K Data set.}
		\begin{adjustbox}{max width=\columnwidth}
			\begin{tabular}{lllllll}
				\toprule
				\textbf{Datasets} &{\textbf{\#Entity}} &{\textbf{\#Relations}} & \textbf{Train} & \textbf{Valid} & \textbf{Test} & \textbf{Period (year)} \\
				\midrule
				Wikidata12K & 12,554 & 24    & 32.5k & 4k    & 4k & 1320 - 2019\\
				YAGO11K & 10,623 & 10    & 16.4k & 2k    & 2k & 1900 - 2017\\
				\bottomrule
			\end{tabular}%
		\end{adjustbox}
		\label{tbl:exp-info}%
	\end{table}%
	\section{Experimental Results}\label{sec:exp}
	This section conducts experiments to evaluate RTGE~\footnote{We will open-source code and data sets upon the publishing of this paper.} and demonstrate its advantages through comparative study.
	
	\begin{table*} 
		\centering
		\caption{Experimental results on Wikidata12K data set. $\downarrow$ means the lower the better. $\uparrow$ means the higher the better.}
		\begin{tabular}{|l|ccc|cc|c|}
			\hline
			Metric & {\textbf{Mean Rank $\downarrow$}} &{\textbf{Mean Rank $\downarrow$}} &{\textbf{Mean Rank $\downarrow$}} &{\textbf{Hits@10(\%) $\uparrow$}} & {\textbf{Hits@10(\%) $\uparrow$}} & \textbf{Hits@1(\%) $\uparrow$} \\
			\hline
			Task  & head  & tail  & relation & head  & tail  & relation \\
			\hline
			Trans-E ~\citep{bordes2013translating} & 740   & 520   & 1.35  & 6     & 11    & 88.4 \\
			TransH ~\citep{wang2014knowledge} & 648   & 423   & 1.4   & 11.8  & 23.7  & 88.1 \\
			DistMult ~\citep{YangYHGD14a} & 635   & 531   & -   & 19.6  & 26.6  & - \\
			ComplEx ~\citep{trouillon2016complex} & 706   & 551   & -   & 11.8  & 23.7  & - \\
			ConvE ~\citep{dettmers2018conve} & 355   & 241   & -   & 25.5  & 33.4  & - \\
			\hline
			HolE ~\citep{nickel2016holographic} & 808   & 734   & 2.23  & 12.3  & 25    & 83.96 \\
			t-TransE ~\citep{jiang2016encoding} & 413   & 283   & 1.97  & 14.5  & 24.5  & 74.2 \\
			HyTE ~\citep{dasgupta2018hyte} & 237   & 179   & \textbf{1.13}  & 25    & 41.6  & \textbf{92.6} \\
			\hline
			RTGE-n ($d=128$)& \textbf{201}   		& \textbf{141}   	& \textbf{1.13}  & \textbf{29.7}  & \textbf{44.6}  & \textbf{92.6} \\
			RTGE-s ($d=128$)& \textbf{210}   		& \textbf{148}   	& \textbf{1.11}  & \textbf{29.5}  & \textbf{43.4}  & \textbf{92.6} \\
			RTGE ($d=128$)& \textbf{183} 	& \textbf{127} 		&\textbf{ 1.09} & \textbf{29.9} & \textbf{44.6} & \textbf{92.8} \\
			RTGE ($d=256$)& \textbf{153} 	& \textbf{91} & \textbf{1.08} & \textbf{32.6} & \textbf{49.7} & \textbf{93.5} \\
			\hline
		\end{tabular}%
		\label{exp:tbl:wy}%
		\vspace{-2mm}
	\end{table*}%
	\subsection{Data Sets with Time-aware Information}
	We conduct extensive experiments on two famous benchmark datasets, the Wikidata12k data set and the YAGO11k data set. 
	YAGO11k is drawn from YAGO3~\cite{mahdisoltani2013yago3} in which some temporally associated facts have meta-facts as (factID, occurSince, start-time), (factID, occurSince, end-time). The total number of time annotated facts in YAGO3 containing both occursSince and occursUntil is 722,494. We choose the top 10 most frequent temporally rich relations of YAGO3 according to the preprocessing method proposed by~\cite{mahdisoltani2013yago3}. To handle sparsity and ensures healthy connectivity within the graph, we recursively delete the edges in the subgraph that contain only one mention containing the entity. Finally, we obtain a purely time-aware graph. We process the Wikidata~\citep{Leblay:2018:DVT:3184558.3191639} dataset according to a similar method of processing YAGO3, and obtain Wikidata12k. Different from YAGO11k, we select the top 24 frequent temporally rich relations for Wikidata12k. 
	
	The time annotations in YAGO11k and Wikidata12k includes the year, month, and day information. Following the setting in HyTE, we drop the month and data information. To distribute the time annotations in the KG uniformly, we club the less frequent year mentions into the same time interval by applying a minimum threshold of 300 triples per interval during construction. For the years with high frequency, we club them into individual intervals. Then we treat timestamps as 61 and 78 different intervals for YAGO and Wikidata, respectively. Statistics of the Wikidata12k and the YAGO11k are summarized in Table~\ref{tbl:exp-info}. 
	
	Although WordNet~\cite{miller1995wordnet} and Freebase~\cite{bollacker2008freebase} are accessible knowledge graph datasets for static knowledge graph research, we do not test on these two datasets. This is because the algorithm proposed in this paper is aimed at the problem of knowledge graph embedding with time information, and the current version of these data sets lacks time information.

	\begin{table*}  
		\centering
		\caption{Experimental results on YAGO11K data set. $\downarrow$ means the lower the better. $\uparrow$ means the higher the better.}
		\begin{tabular}{|l|ccc|cc|c|}
			\hline
			Metric & 
			{\textbf{Mean Rank $\downarrow$}} &
			{\textbf{Mean Rank $\downarrow$}} &
			{\textbf{Mean Rank $\downarrow$}} &
			{\textbf{Hits@10(\%) $\uparrow$}} & 
			{\textbf{Hits@10(\%) $\uparrow$}} &
			\textbf{Hits@1(\%) $\uparrow$} \\
			\hline
			Task  & head  & tail  & relation & head  & tail  & relation \\
			\hline
			Trans-E ~\citep{bordes2013translating} & 2020  & 504   & 1.7   & 1.2   & 4.4   & 78.4 \\
			TransH ~\citep{wang2014knowledge} & 1808  & 354   & 1.53  & 1.5   & 5.8   & 76.1 \\
			DistMult ~\citep{YangYHGD14a} & 1550   & 616   & -   & 17.3  & 31.4  & - \\
			ComplEx ~\citep{trouillon2016complex} & 1758   & 603   & -   & 19.2  & 35.3  & - \\
			ConvE ~\citep{dettmers2018conve} & 1464   & 702   & -   & 19.2  & 33.5  & - \\
			\hline
			HolE ~\citep{nickel2016holographic} & 1953  & 1828  & 2.57  & 13.7  & 29.4  & 69.3 \\
			t-TransE ~\citep{jiang2016encoding} & 1692  & 292   & 1.66  & 1.3   & 6.2   & 75.5 \\
			HyTE ~\citep{dasgupta2018hyte} & 1069  & 107   & 1.23  & 16    & 38.4  & 81.2 \\
			\hline
			RTGE-n ($d=128$)& \textbf{854}   	& 113   &\textbf{ 1.22}  	& \textbf{20.52} 	& \textbf{40.9} & \textbf{83.9} \\
			RTGE-s ($d=128$)& \textbf{891}   	& 118   & 1.24  	& \textbf{20.1}  	& \textbf{39.1}  & \textbf{82.8} \\
			RTGE ($d=128$)& \textbf{799} 		& 110 	& \textbf{1.15} 		& \textbf{20.1}  	& \textbf{40.9} 	& \textbf{88.2} \\
			RTGE ($d=256$)& \textbf{725} 	& \textbf{105} 	& \textbf{1.11} & \textbf{21.4} & \textbf{41.5} & \textbf{88.9} \\
			\hline
		\end{tabular}%
		\label{exp:tbl:wy2}%
		\vspace{-2mm}
	\end{table*}%
	
	\subsection{Entity, Relation, and Temporal Scoping Prediction Tasks}
	To verify the performance of RTGE, we conduct four types of tasks on each time-aware dataset (including head entity prediction, tail entity prediction, relationship prediction, and temporal scoping prediction). The settings for these tasks are as described in the problem statement section. For a given test triplet with missing entities or relationships, we use formula (1) to calculate the loss of all potential entities or relationships with the test triplet under this formula. We sort the potential entities or relationships in ascending order according to the loss value. For evaluation, we select the ranking of real entities or relationships corresponding to the triples. For a given test triple with missing time, we project the entities and relationships in the triple to all potential time hyperplanes. We calculate the loss of all potential time and test triples under this formula according to formula (1), and sort the potential time according to the loss value. For evaluation, we select the real-time ranking corresponding to the triple. In these tasks, we follow a time agnostic negative sampling procedure to generate negative samples. For instance, with a given tail and head query term, we randomly replace a tail or head entity such that newly generated triple is not observed.

	\subsection{Baselines}
	We compare RTGE with a set of state-of-the-art graph embedding algorithms. 
	\begin{itemize}
		\item Firstly, considering that our RTGE is a kind of translation-style approach, we compare with Trans-E and TransH, which are two translation-based graph embedding algorithm. 
		
		\item Secondly, noting that the outstanding performance of graph neural network algorithms in graph embedding has received increasing attention from researchers in recent years, we include DistMult, ComplEx, and ConvE as baselines for comparison. 
		
		\item Thirdly, considering that we have motivated the study by using time-aware information, whereas HolE, t-TransE, and HyTE are known as an effective time-aware graph embedding algorithm for entity or relation prediction tasks, we include they as baselines for comparison. 
		
		\item Regarding our proposed RTGE, to further investigate the impact of the negative relation sampling and the temporal smoothness to the overall performance, we denote by RTGE-s a reduction of RTGE only using temporal smoothness and randomly selecting the negative entity sampling. We denote by RTGE-n a reduction of RTGE only using the negative relation sampling and learning all the hyperplanes independently.    
	\end{itemize}

	\begin{figure*}[t!]
		\centering
		\subfigure[Wikidata12K.]
		{\includegraphics[width=0.48\textwidth]{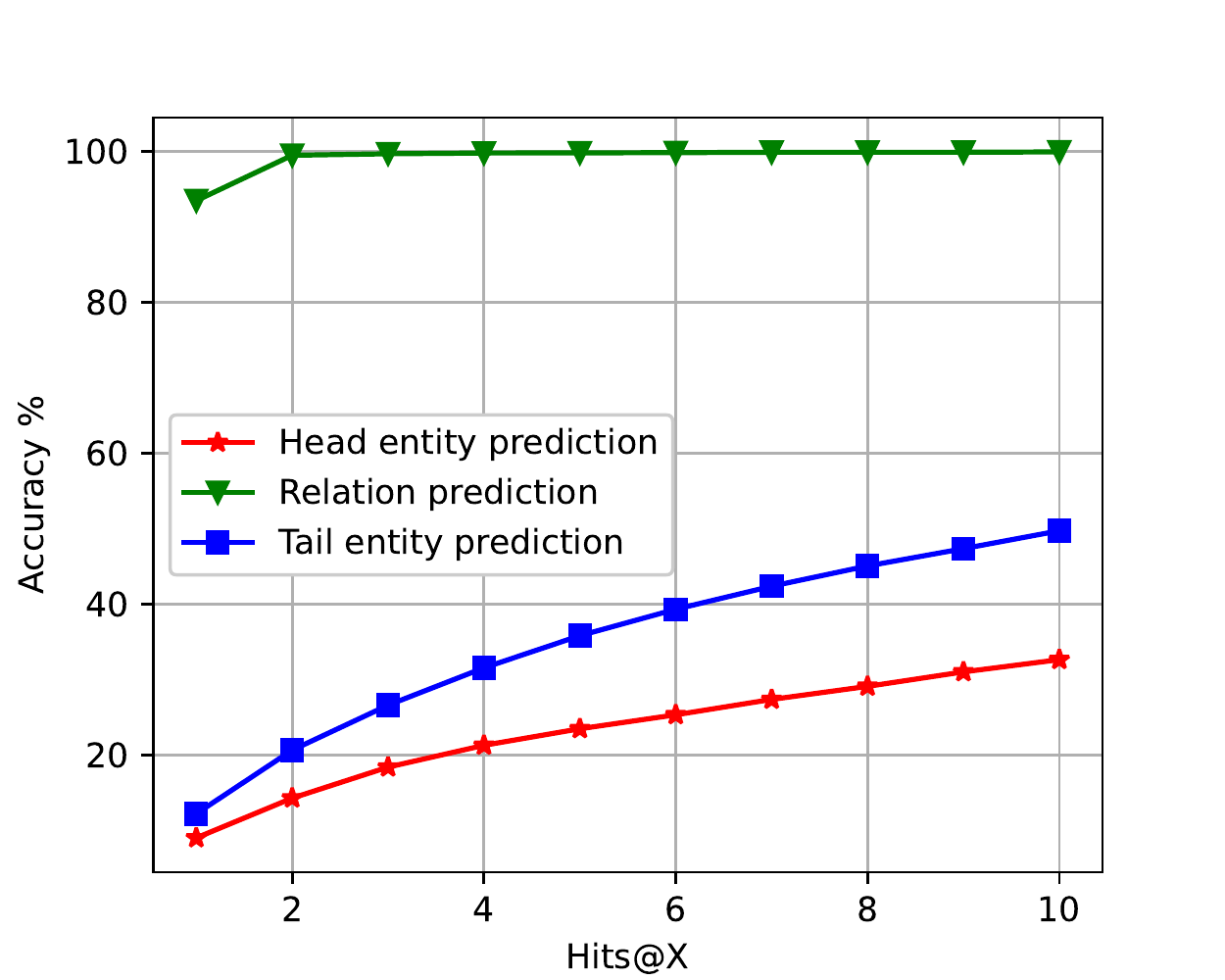}\label{exp:fig:htis:wiki_data}}
		\subfigure[YAGO11K.]
		{\includegraphics[width=0.48\textwidth]{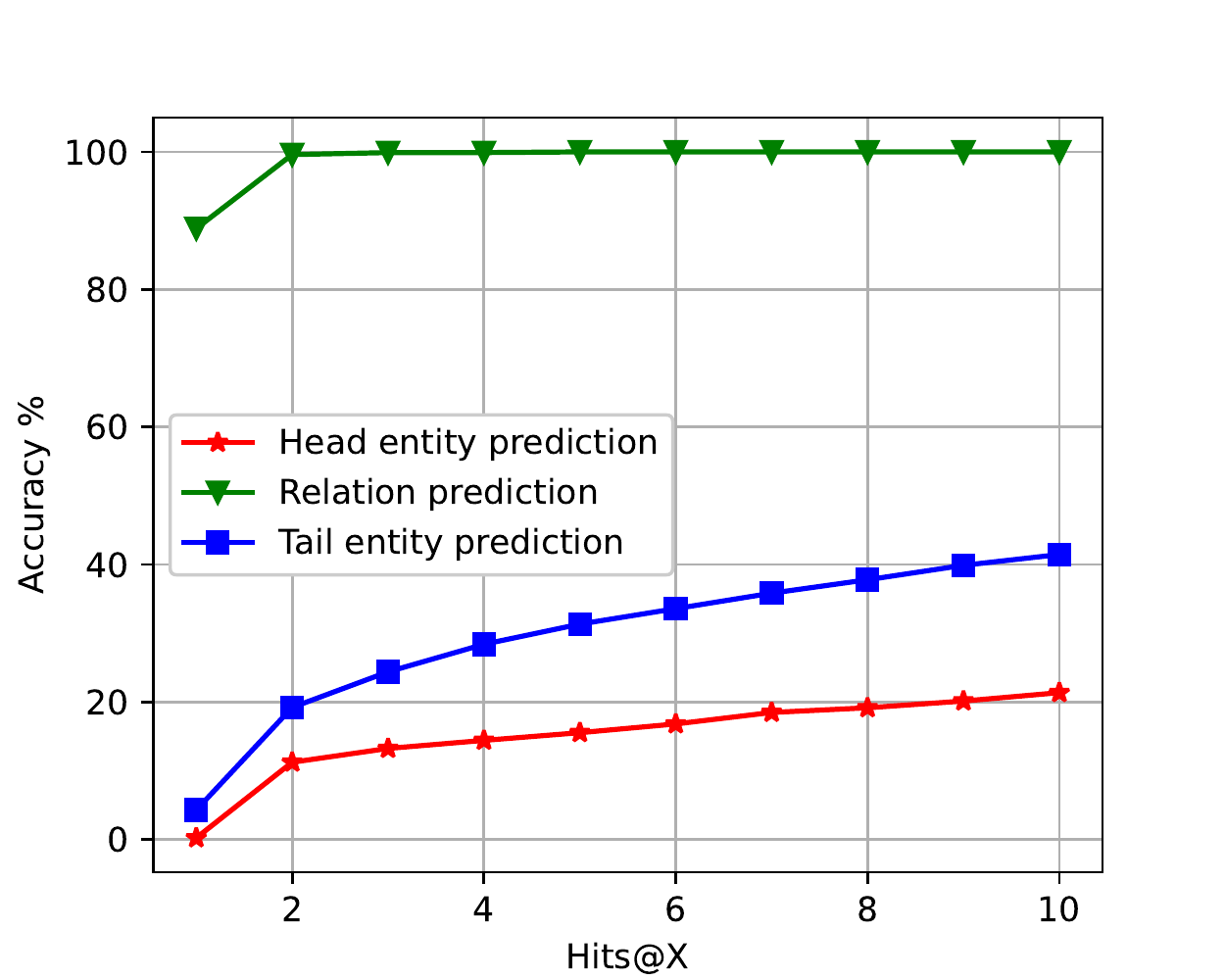}\label{exp:fig:htis:yago}}
		\caption{Experimental results of RTGE on Wikidata12K and YAGO11K data sets with respect to different Hits@X.}\label{exp:fig:hits}
		\vspace{-5mm}
	\end{figure*}
	
	We use similar evaluation metrics as traditional knowledge graph embedding methods for time-aware knowledge graph embedding methods. RTGE utilizes equation (\ref{eq:loss-l}) to calculate the loss corresponding to triples formed by each potential head entity and the observed tail entity and relationship. We calculate and record the ranking of the loss for the real head entity after sorting all losses. Then, we report the mean rank, Hits@1, Hits@2,$\dots$, and Hits@10. Other prediction tasks use the same method to evaluate the performance of the algorithms.
	
	\subsection{Qualitative Results}
	Table~\ref{exp:tbl:wy} and Table~\ref{exp:tbl:wy2} reports the experimental results on Wikidata12K and YAGO11K data sets.  We compare and analyze the performance differences of the proposed RTGE and benchmark algorithms from the following five aspects. In order to show the performance of RTGE on two benchmark datasets in more detail, we report the experimental results of RTGE on Wikidata12K and YAGO11K data sets concerning different Hits@X in Figure.~\ref{exp:fig:hits}. 
	It can be seen from the figure that when $X$ is smaller, the algorithm is less effective in the entity prediction task. In the entity prediction task, the smaller $X$ is not enough to reflect the overall performance of the algorithm. As $X$ gradually increases to 10, the effect of the algorithm in the entity prediction task is significantly improved. A larger $X$ can better reflect the effect of the algorithm.
	In contrast, in the relationship prediction task, when $X$ is larger, the accuracy of the algorithm always approaches 100\%. This is not conducive to the performance of the comparison algorithm. Therefore, a smaller $X$ is more suitable for the relationship prediction task.
	
	\begin{figure}[]
		\centering
		{\includegraphics[width=0.48\textwidth]{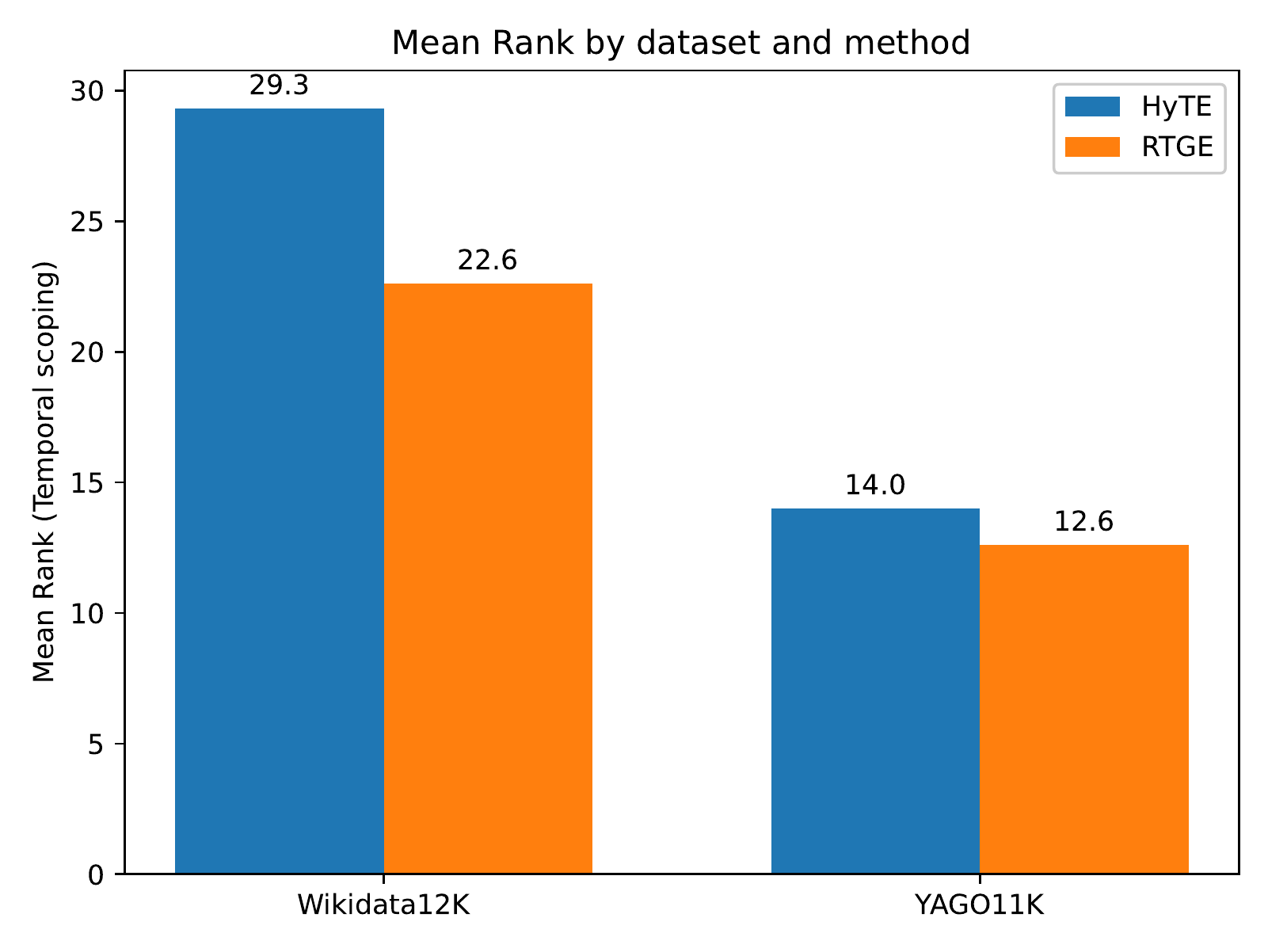}}
		\caption{The predicted Mean Rank (lower the better) for temporal scoping. }\label{exp:fig:time_scope_rank}
		\vspace{-5mm}
	\end{figure}
	\subsubsection{Comparative Analysis with Translation-based algorithms}
	From the Table~\ref{exp:tbl:wy} and Table~\ref{exp:tbl:wy2}, we can observe that the average performances of Trans-E and TransH on all the three tasks are worse than the performance of RTGE. This is because Trans-E and TransH only learn one embedding for each entity or relationship at different time intervals. However, as time changes, new entities or relationships are added to the knowledge graph, and old entities or relationships disappear. This means that the structure of the knowledge graph is constantly changing at different time intervals. It is insufficient to rely on a fixed embedding to express entities or relationships at different times in time.
	
	\subsubsection{Comparative Analysis with Graph Neural Network-based algorithms}
	Table~\ref{exp:tbl:wy} and Table~\ref{exp:tbl:wy2} show that the three graph-based neural network algorithms DistMult, ComplEx, and ConvE have significantly better execution results than the translation-based algorithms Trans-E and TransH. However, due to the lack of modeling of time information, even if DistMult, ComplEx, and ConvE use more complex neural network models, the effect is still no better than RTGE. These results indicate that the static knowledge graph embedding algorithm has inherent flaws when processing time-aware knowledge graphs.
	
	\subsubsection{Comparative Analysis with Time-aware Graph Embedding algorithms}
	Despite t-TransE and HyTE both take into account the time factor and try to use temporal information during model transformation or projected-time translation, their performance still does not exceed RTGE-s and RTGE. This is because RTGE not only considers the graph structure information at different timestamps but also consider the association between graph structures between adjacent timestamps. Due to this advantage, RTGE learns the evolution of time-aware graphs over time more accurately, thus predicting entities or relationships more accurately at a future moment.
	
	\begin{figure}[t!]
		\centering
		{\includegraphics[width=0.48\textwidth]{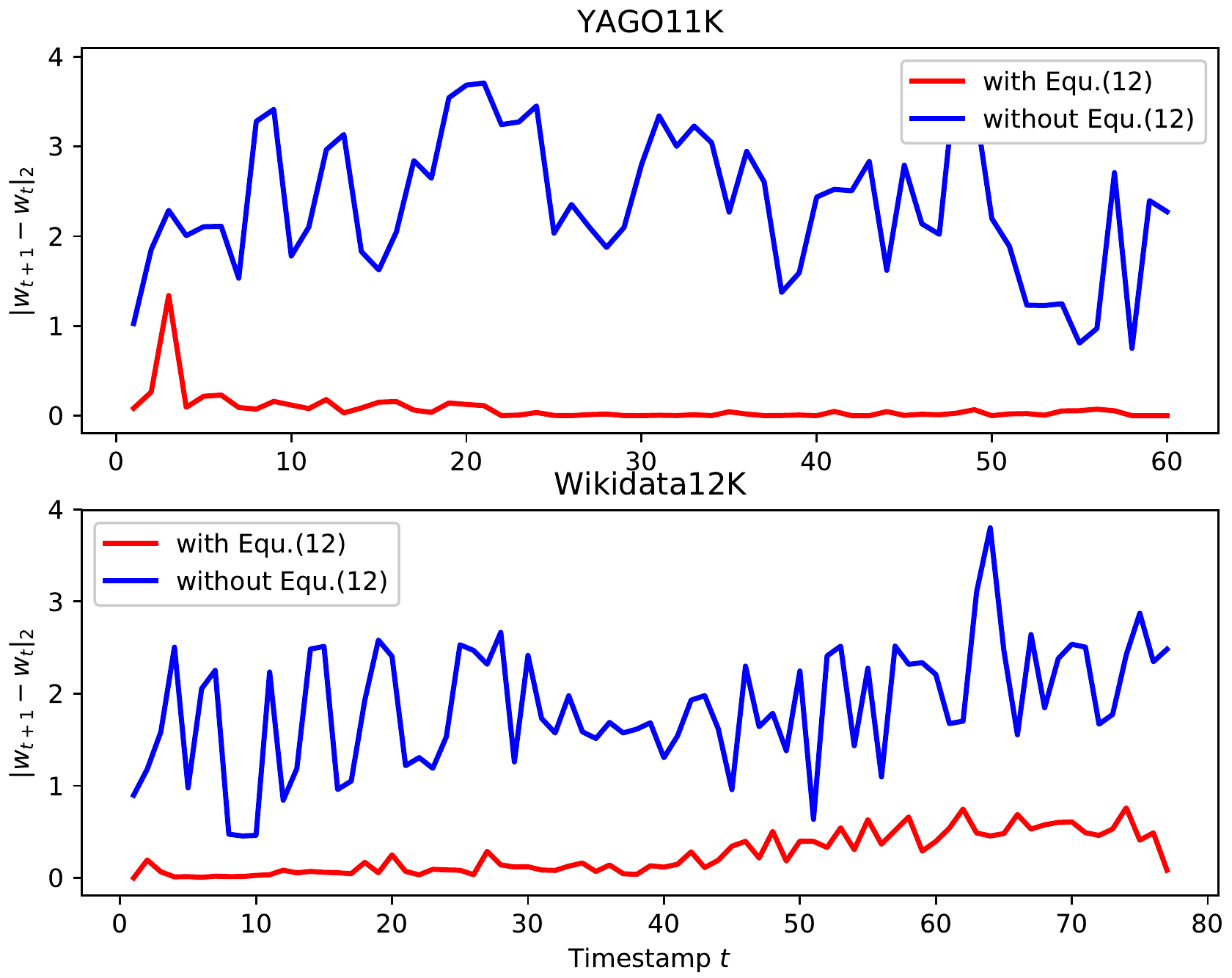}}
		\caption{Statistics of $w_t$ based on time embeddings obtained after training RTGE with / without equation (\ref{eq:obj:smooth}). }\label{exp:fig:l2_wt}
		\vspace{-5mm}
	\end{figure}
	
	\begin{figure*}[t!]
		\centering
		\subfigure[Wikidata12K (RTGE).]
		{\includegraphics[width=0.48\textwidth]{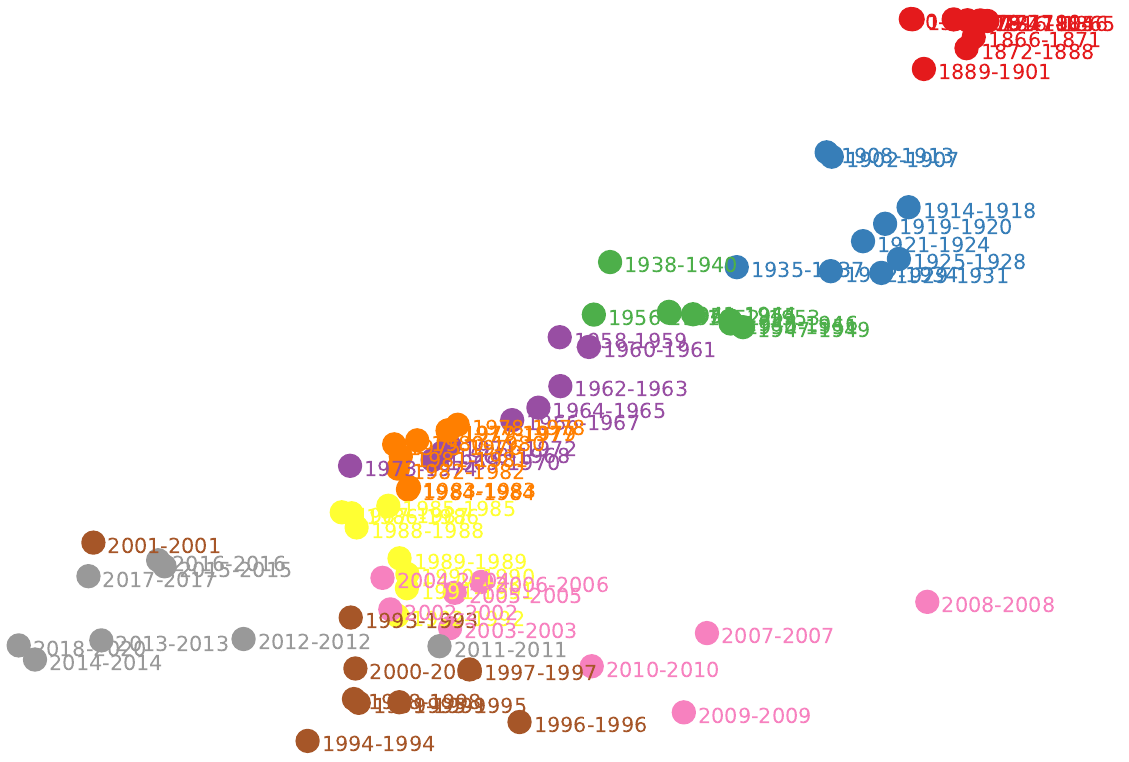}\label{exp:fig:param:wiki_data_e}}
		\subfigure[YAGO11K (RTGE).]
		{\includegraphics[width=0.48\textwidth]{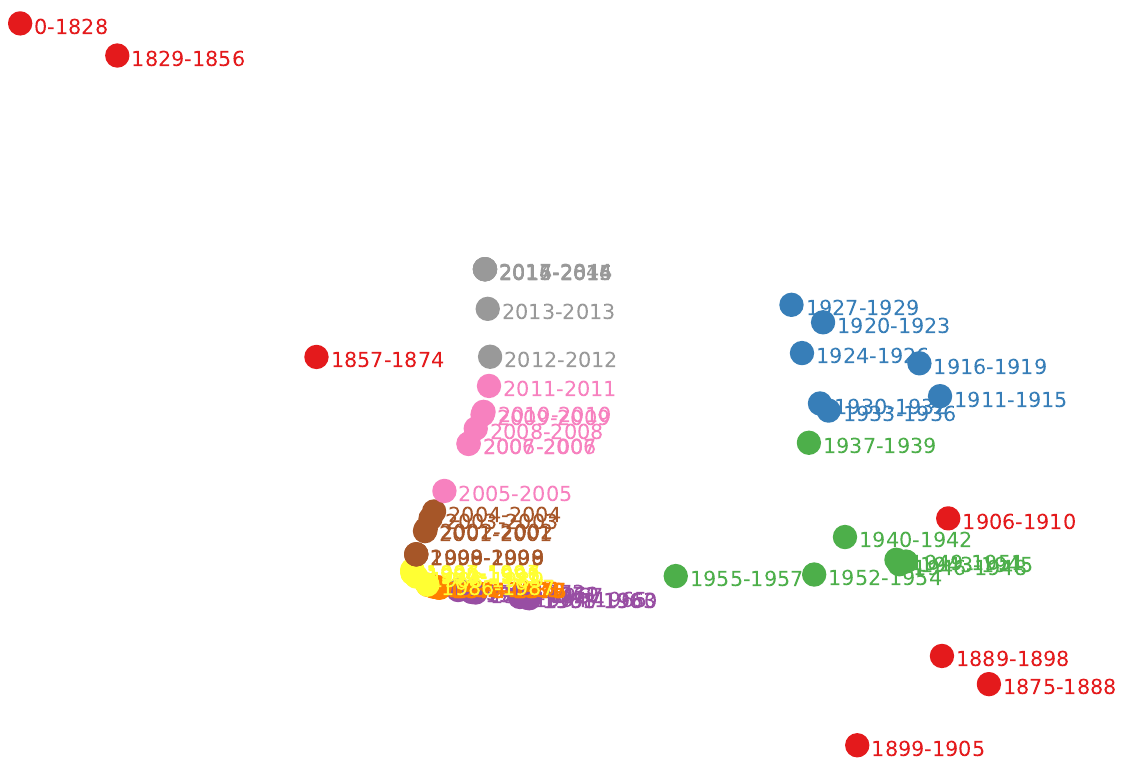}\label{exp:fig:param:yago_e}}
		\subfigure[Wikidata12K (HyTE).]
		{\includegraphics[width=0.48\textwidth]{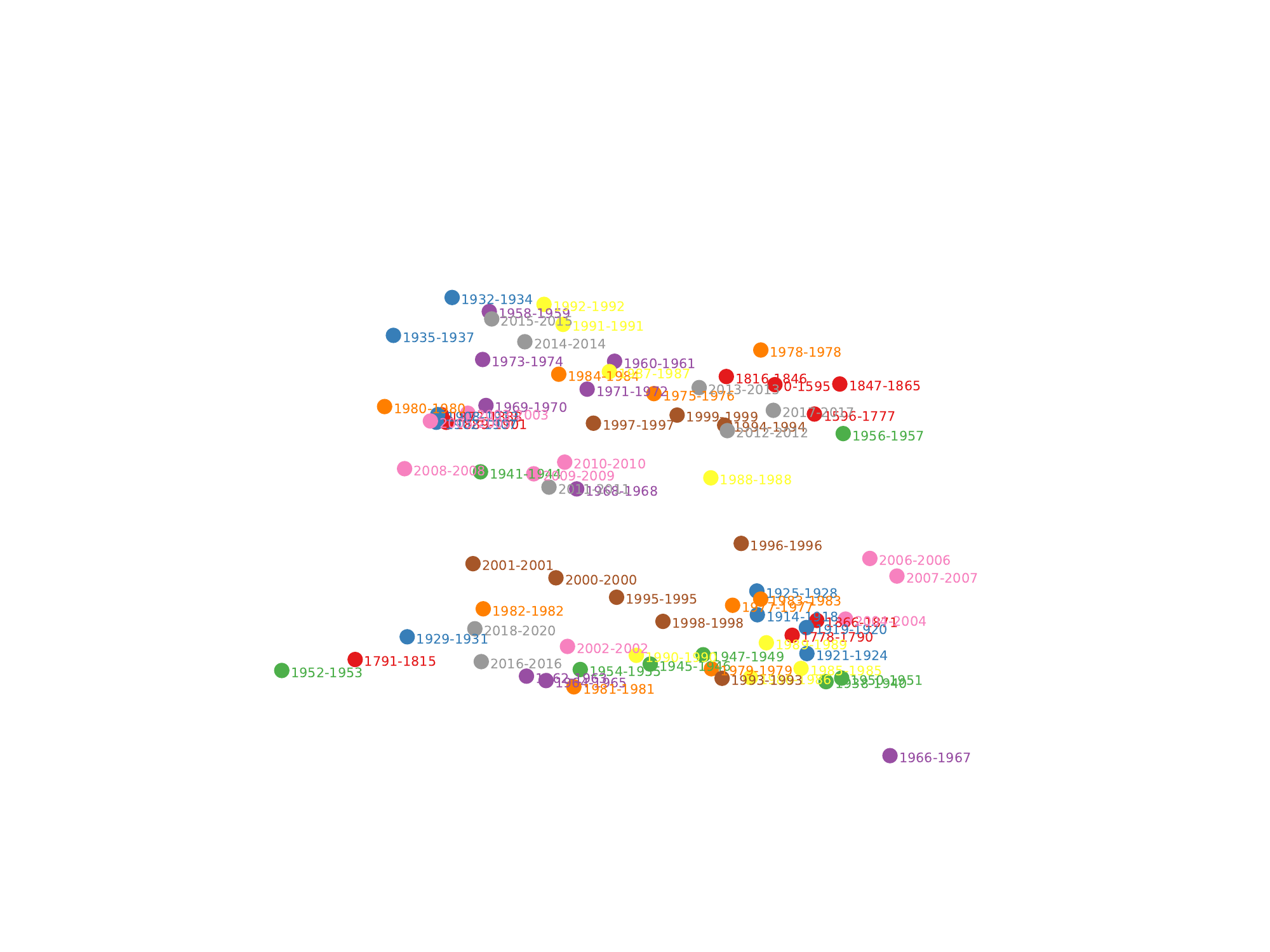}\label{exp:fig:param:wiki_data_e:hyte}}
		\subfigure[YAGO11K (HyTE).]
		{\includegraphics[width=0.48\textwidth]{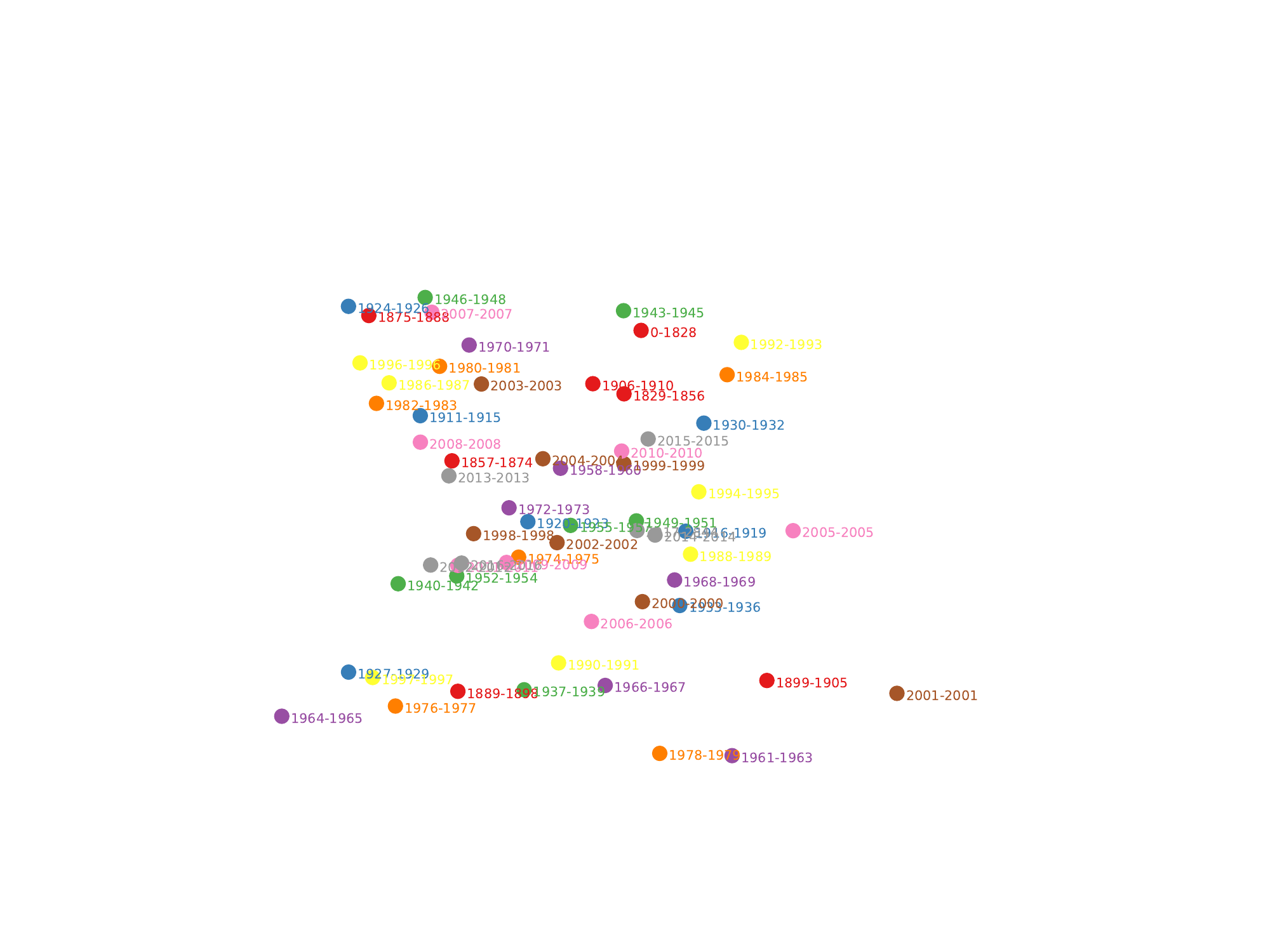}\label{exp:fig:param:yago_e:hyte}}
		\caption{The figure illustrates 2-d PCA projection of time embeddings which are obtained after training RTGE and HyTE for temporal scoping task. Dots of the same color indicate adjacent time embeddings.}\label{exp:fig:time:e}
		\vspace{-5mm}
	\end{figure*}
	\subsubsection{Verification on Temporal Smoothness and Task-Oriented Negative Sampling}
	By comparing RTGE-s and HyTE in Table~\ref{exp:tbl:wy} and Table~\ref{exp:tbl:wy2}, we can verify the effectiveness of temporal smoothness.
	From the experimental results in the two tables, it can be seen that compared to HyTE, RTGE-s achieved a better result in most test groups.
	This is because, compared to HyTE, RTGE-s introduces temporal smoothness terms in the embedding model by constraining the variation between hyperplanes at adjacent timestamps. In this way, RTGE-s avoids the challenge of the anomaly data to the model and indirectly improves the robustness of the graph embedding algorithm. Therefore, RTGE-s achieve better performance than HyTE.
	
	By comparing RTGE-n and HyTE in Table~\ref{exp:tbl:wy} and Table~\ref{exp:tbl:wy2}, we can verify the effectiveness of task-oriented negative sampling. From the figues, we can find that RTGE-n not only outperforms HyTE in relational prediction tasks but also outperforms HyTE in entity prediction tasks. This is because, compared to HyTE, RTGE-n introduces negative sampling triplets for the relationship in the embedding model, which significantly improves the discriminative ability of the model for different relationships, and it further enhances the performance of the embedding model in the entity prediction task.

	\begin{figure*}[t!]
		\centering
		\subfigure[Sensitivity study on $m$.]
		{\includegraphics[width=0.8\textwidth]{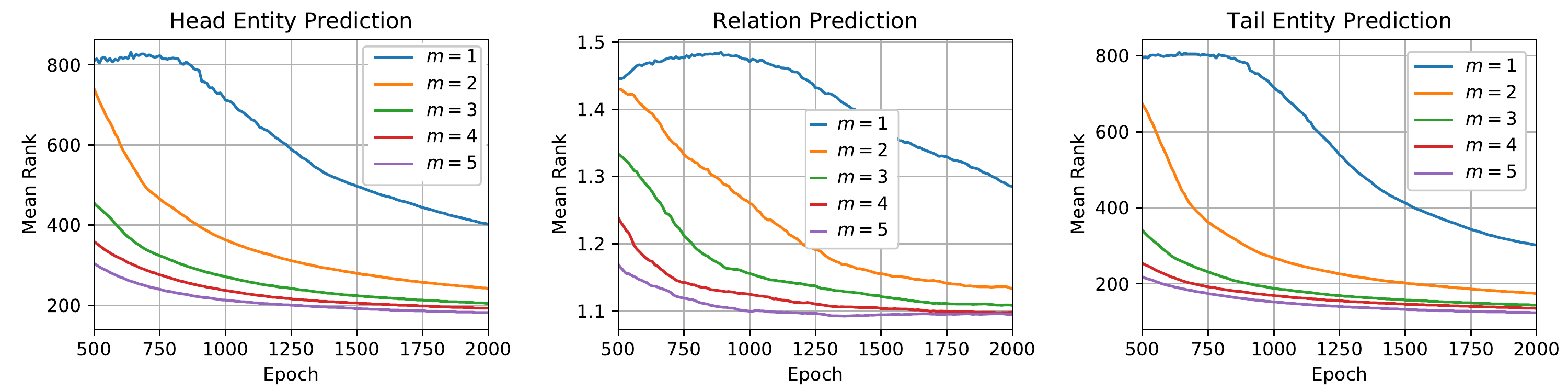}\label{exp:fig:param:m}}
		\subfigure[Sensitivity study on $\alpha$.]
		{\includegraphics[width=0.8\textwidth]{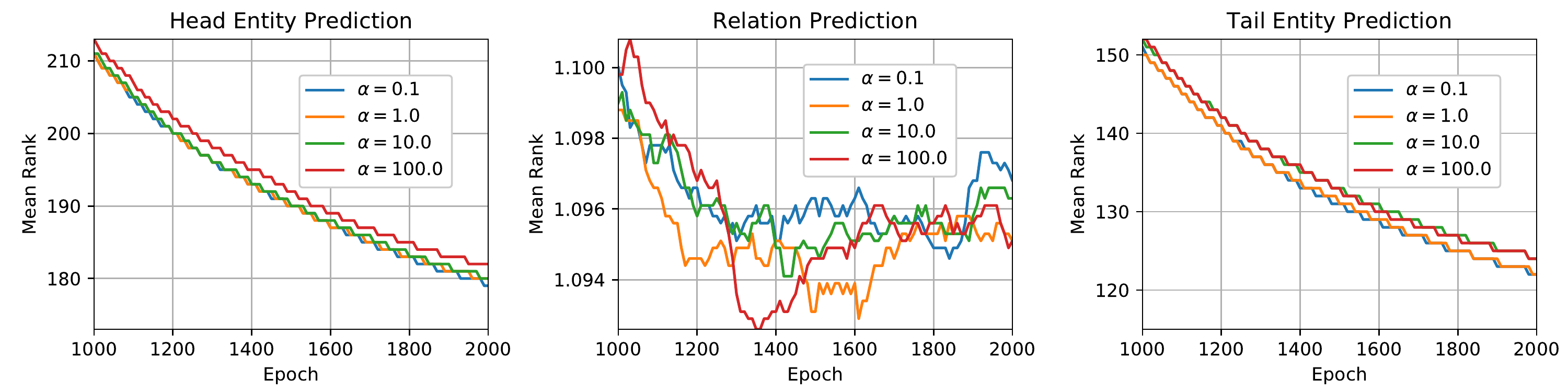}\label{exp:fig:param:alpha}}
		\subfigure[Sensitivity study on $\beta$.]
		{\includegraphics[width=0.8\textwidth]{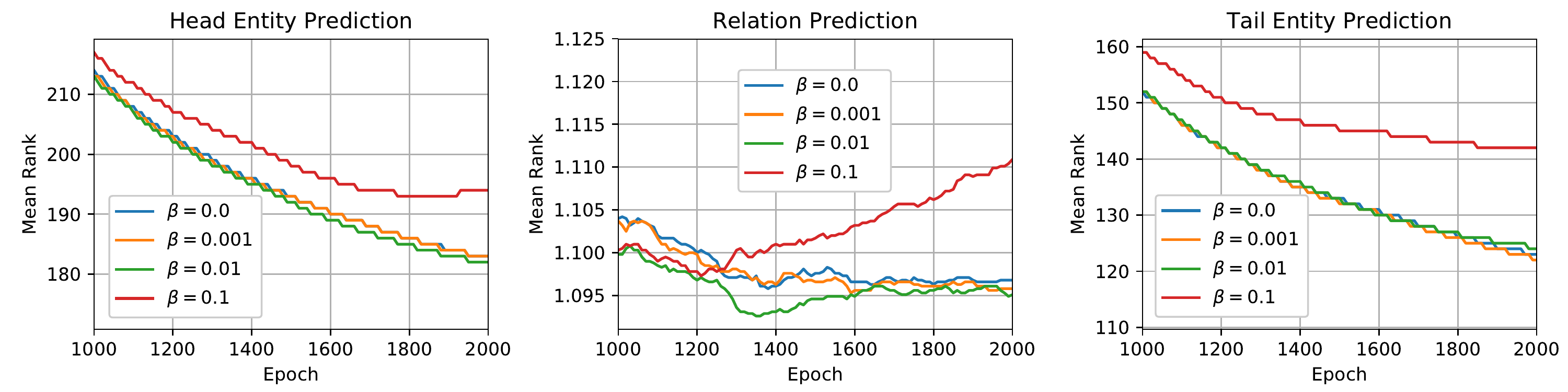}\label{exp:fig:param:beta}}
		\subfigure[Sensitivity study on $\gamma$.]
		{\includegraphics[width=0.8\textwidth]{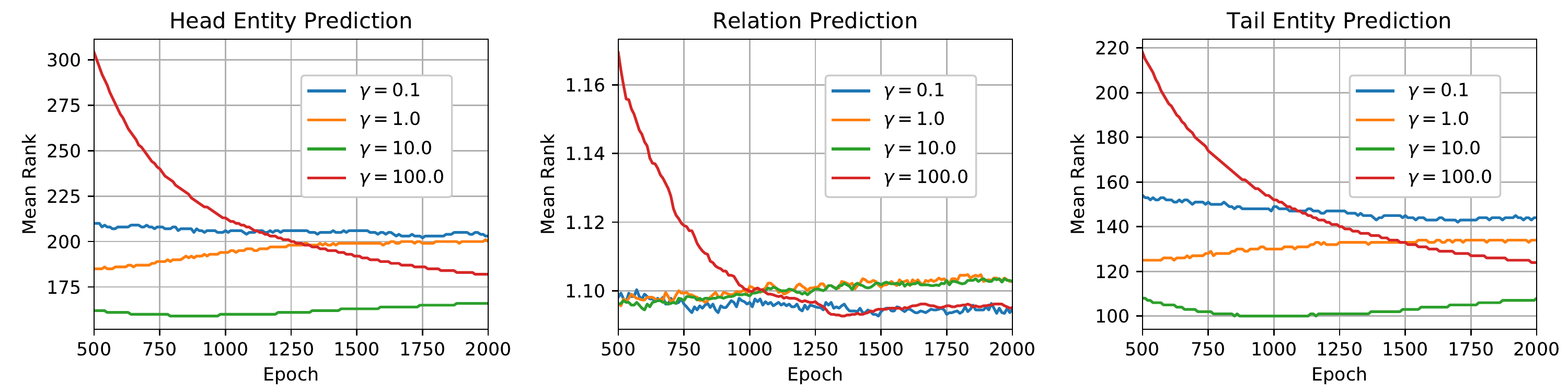}\label{exp:fig:param:gamma}}
		\caption{Sensitivity study of $m$, $\alpha$, $\beta$ and $\gamma$ on Wikidata12K.}\label{exp:fig:param}
		\vspace{-5mm}
	\end{figure*}
\begin{figure*}[t!]
\centering 
\subfigure[Wikidata12K.]
{\includegraphics[width=0.8\textwidth]{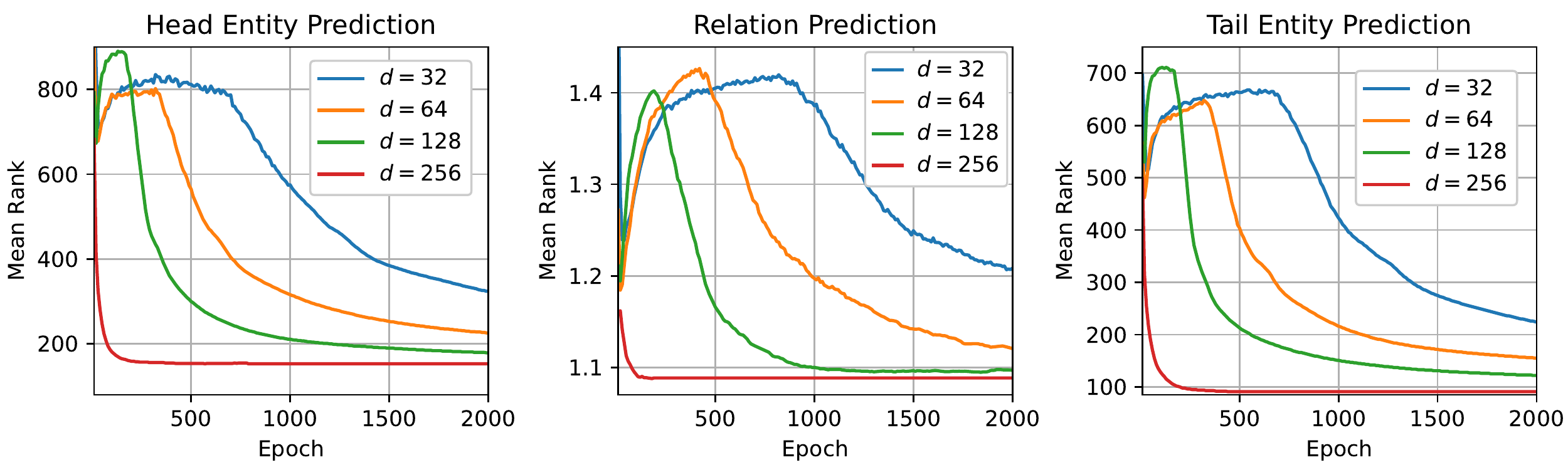}\label{exp:fig:param:wiki_data_dim}}
\subfigure[YAGO11K.]
{\includegraphics[width=0.8\textwidth]{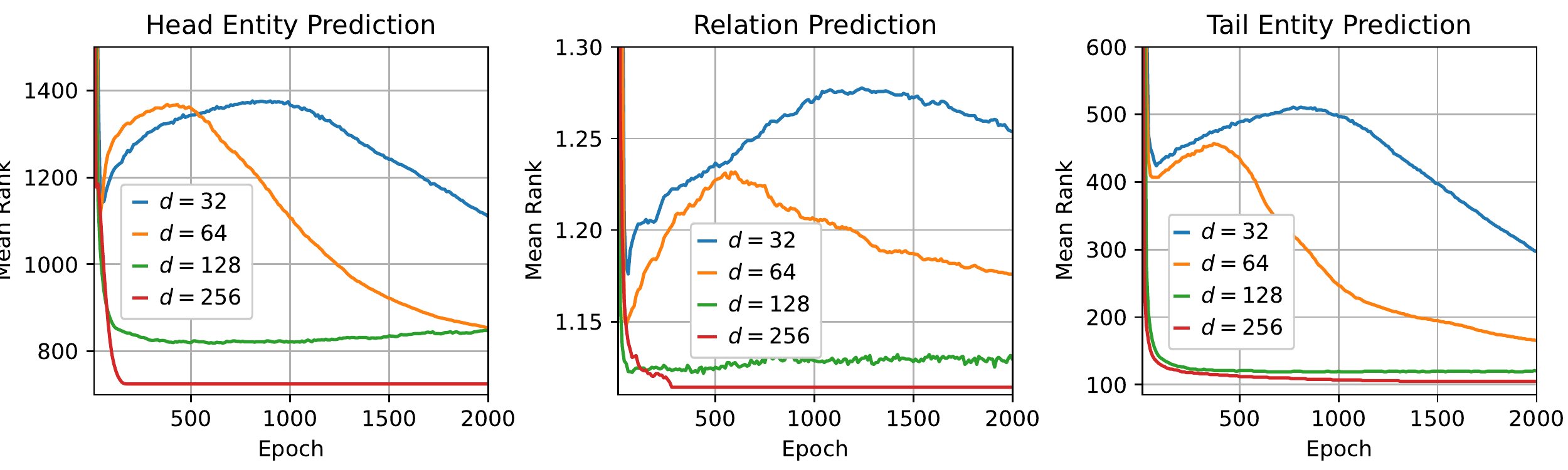}\label{exp:fig:param:yago_dim}}
\caption{Sensitivity study of $d$.}\label{exp:fig:param:dim}
\vspace{-4mm}
\end{figure*}
	
	\subsubsection{Temporal Scoping Prediction Result}
	Fig.~\ref{exp:fig:time_scope_rank} shows the experimental results of temporal scoping prediction. It can be seen from the figure that the predicted mean rank for temporal scoping of RTGE is lower than in HyTE. This is because HyTE does not use a timing smoothing mechanism, which is not conducive to learning time embedding to maintain timing information. Hence, HyTE is not conducive to temporal scoping for accurately predicting events. On the contrary, RTGE introduces a temporal smoothness mechanism, which can well avoid this problem encountered in HyTE. Fig.~\ref{exp:fig:l2_wt} shows the $\|w_{t+1}-w_t\|_2$ based on time embeddings which are obtained after training RTGE with or without equation (\ref{eq:obj:smooth}). It can be seen from the figure that in the $w_t$'s learned by RTGE with equation (\ref{eq:obj:smooth}), the change of $w_t$ at the neighboring moment is relatively smooth, and the change range is relatively small, which is benefited from the timing smoothing mechanism of RTGE. However, in the $w_t$'s learned by RTGE without equation (\ref{eq:obj:smooth}), the variation range of $w_t$ at the neighboring moment is large and irregular, which is not conducive to maintaining the consistency of the embedded timing.
	
	In order to show the time embedding learned by RTGE and HyTE more intuitively, Fig.~\ref{exp:fig:time:e} shows the 2-d PCA projection of time embeddings, which are obtained after training RTGE for the temporal scoping task.  We observe that the time representation after training RTGE is forming natural clusters in chronological order. However, the time representation after training HyTE is more evenly distributed in the figure. Only a small number of adjacent time representations come together. This indirectly verifies that the time embedding learned by the RTGE model can effectively retain the time sequence information in the time-aware knowledge graph.

	\subsection{Parameter Tuning}
	This section shows the results of RTGE's parameter tuning experiments, including the trade-off parameter $\beta$, the number of negative sampling triples $m$, the trade-off parameter $\alpha$, $\xi$, the learning rate $\psi$, the margin $\gamma$, embedding dimension $d$. In our experiments, $\xi$, and $\psi$ are empirically set to 1 and 0.0001 on data sets. In order to shorten the parameter tuning time, we set $d=128$ in the adjustment experiment of parameter $m$, $\alpha$,$\beta$ and $\gamma$.
	
	\subsubsection{Sensitivity Study on $m$}
	Fig.~\ref{exp:fig:param:m} shows the sensitivity study result of $m$. 
	As can be seen from the figure, when $m$ is small, the performance of RTGE is poor. For example, on Wikidata12K dataset, when $m = 1$, the mean rank of RTGE on the head entity prediction task is 400. As $m$ increases, the performance of RTGE continues to improve. When $m = 5$, the mean rank of RTGE on the head entity prediction task is lower than 200. This result indicates that the appropriate increase in the number of negative sampling triples will help to improve the discriminative ability of RTGE. Besides, it can be observed that with the continuous increase of $m$, the improvement of RTGE performance is decreasing. This result indicates that too large $m$ cannot continue to significantly improve the performance of RTGE. Too large $m$ will bring more training samples, which will significantly increase the time required for training. Given these observations, we set $m = 5$.
	
	\subsubsection{Sensitivity Study on $\alpha$}
	Fig.~\ref{exp:fig:param:alpha} provides the sensitivity study result of $\alpha$. 
	It can be seen from the figure that the sensitivity of $\alpha$ on different data sets and different tasks varies greatly. For example, in the test of the Wikidata12K dataset, a smaller $\alpha$ is beneficial to improve the performance of RTGE in the entity prediction task. In contrast, in relation prediction, RTGE tends to choose a larger $\alpha$.
	In view of these results, we choose different $\alpha$'s for RTGE on different datasets and different tasks. For example, on the Wikidata12K dataset, we choose $\alpha = 0.1$ for the entity prediction task, and $\alpha = 100$ for the relation prediction task. On the YAGO11K dataset, we choose $\alpha = 100$ for all tasks.
	
	\subsubsection{Sensitivity Study on $\beta$}
	Fig.~\ref{exp:fig:param:beta} reports the potential impacts imposed by different numbers of trade-off parameter, $\beta$, to the mean rank of RTGE. From Fig.~\ref{exp:fig:param:beta}, we can observe that when the epoch is small, the mean rank of RTGE on the three tasks is more significant. As epoch increases, the mean rank of RTGE gradually decreased and stabilized in most test groups. Moreover, when $\beta$ is less than or equal to 0.01, the mean rank of RTGE can get the smallest value at the maximum epoch. These results indicate that an oversized $\beta$ increases the performance of RTGE on Wikidata12K, which is not conducive to prediction. Therefore, we set the number of the trade-off parameter $\beta$ to 0.01 for Wikidata12K.

	\subsubsection{Sensitivity Study on $\gamma$} 
	To analyze the effect of different values of $\gamma$ on the prediction performance of RTGE, the sensitivity study experiment is conducted on entity and relation prediction tasks where $\gamma$ is selected from 0.1 to 100. Fig.~\ref{exp:fig:param:gamma} shows the experimental results. We observe that, on the Wikidata12K dataset, the entity prediction task tends to choose $\gamma=10$.  And $\gamma=10$ or $\gamma=0.1$ is more suitable for the relation prediction task.
	Different the result on Wikidata12K, $\gamma=100$ is more suitable for the head entity prediction task on YAGO11K. Furthermore, $\gamma=0.1$ is the best choice for the tail entity and relation prediction tasks on YAGO11K.
	
	\subsubsection{Sensitivity Study on $d$}
	To find a suitable dimension of the embedding, we tested the performance of RTGE on a benchmark dataset using different dimensions of embeddings. The experimental results are shown in Fig.~\ref{exp:fig:param:dim}.
	From the figure, we can find an obvious rule. As the epoch increases, the mean rank of the RTGE embedded with different dimensions is decreasing. Higher-dimensional embeddings can make RTGE's mean rank drop faster than lower-dimensional embeddings, and they can always help to reach the lowest mean rank. In particular, the performance of RTGE in multiple tasks of two data sets is optimal when $d=256$. This is because higher-dimensional embeddings can describe more detailed graph structure information than lower-dimensional embeddings. Therefore, higher-dimensional embeddings can achieve the best result. However, as the dimensions continue to increase, the training and testing of the algorithm bring more time loss, which can be verified from the section~\ref{sec:model}.

	\section{Conclusion}\label{sec:conclusion}
	We propose a robustly time-aware graph embedding method. In contrast with other state-of-the-art methods, the main characteristics of our approach are two folds.
	Firstly, RTGE proposes incorporating temporal smoothness into the learning framework of embedding. By taking into account the temporal smoothness between hyperplanes of adjacent time steps, RTGE can more efficiently obtain the evolution of KGs and avoid the interpretation of the wrong conclusion.
	Secondly, RTGE proposes a task-oriented sampling strategy for different tasks that can dynamically adjust the negative sampling ratio of entities and relationships for the characteristics of different tasks. The expanded negative sampling strategy provides RTGE with a stronger adaptive ability to separate tasks. 
	Experiments on Wikidata12K data set and YAGO11K data set verify the superiority of RTGE over other state-of-the-art baseline methods. Although the algorithm proposed in this paper has achieved good results on the benchmark data set, RTGE still uses the TransH-like embedding principle to learn time-based hyperplanes at independent intervals. Compared to TransH, the embedding principles mentioned in TransF and ConvE have better performance. If the embedding principles like TransF and ConvE are introduced into RTGE, it is possible to improve the performance of the algorithm further. This is our future work.

	\ifCLASSOPTIONcaptionsoff
	\newpage
	\fi

	
	
	%
	\bibliographystyle{IEEEtran}
	\bibliography{IEEEabrv,mybib}

\end{document}